\documentclass[letterpaper]{article} % DO NOT CHANGE THIS
\usepackage{aaai24}  % DO NOT CHANGE THIS
\usepackage{times}  % DO NOT CHANGE THIS
\usepackage{helvet}  % DO NOT CHANGE THIS
\usepackage{courier}  % DO NOT CHANGE THIS
\usepackage[hyphens]{url}  % DO NOT CHANGE THIS
\usepackage{graphicx} % DO NOT CHANGE THIS
\usepackage{natbib}  % DO NOT CHANGE THIS AND DO NOT ADD ANY OPTIONS TO IT
\usepackage{caption} % DO NOT CHANGE THIS AND DO NOT ADD ANY OPTIONS TO IT
\usepackage[utf8]{inputenc} % allow utf-8 input
\usepackage[T1]{fontenc}    % use 8-bit T1 fonts
\usepackage{hyperref}       % hyperlinks
\usepackage{url}            % simple URL typesetting
\usepackage{booktabs}       % professional-quality tables
\usepackage{amsfonts}       % blackboard math symbols
\usepackage{nicefrac}       % compact symbols for 1/2, etc.
\usepackage{microtype}      % microtypography
\usepackage{xcolor}         % colors
\usepackage{subcaption} 

\usepackage{amsmath}
\usepackage{amssymb}
\usepackage{mathtools}
\usepackage{amsthm}
\usepackage{multirow}
\usepackage{mathrsfs}
\usepackage{chngcntr}
\usepackage{lipsum}
\usepackage{mathtools, nccmath}
\usepackage{xcolor}
\usepackage{array}
\usepackage{nameref}

\usepackage{blkarray}
\usepackage{bm}

\usepackage{algorithm}% http://ctan.org/pkg/algorithms
\usepackage{algpseudocode}% http://ctan.org/pkg/algorithmicx

\usepackage{abbreviation_definitions}
\newcommand{\Algname}{\texttt{UTE}}
\newcommand{\TEE}{\texttt{$\epsilon z$-Greedy}}
\newcommand{\TempoRL}{\texttt{TempoRL}}
\newcommand{\DAR}{\texttt{DAR}}
\newcommand{\DQN}{\texttt{DQN}}
\newcommand{\DDQN}{\texttt{DDQN}}
\newcommand{\BootstrappedDQN}{\texttt{B-DQN}}

\newcommand{\Full}{\texttt{Full Ensemble}}

\newcommand{\rc}{extension length} % corresponds to skip in skip length
\newcommand{\repetition}{extension} % corresponds to skip in skip policy or skip Q value
\newcommand{\riskparameter}{uncertainty parameter}

\newcommand{\Q}{Q}
\newcommand{\actionpolicy}{\pi_a}
\newcommand{\extensionpolicy}{\pi_e}
\newcommand{\policyoveroption}{\pi_{\omega}}
\newcommand{\behaviorQ}{Q^{\policyoveroption{}}}
\newcommand{\skipQ}{\Tilde{Q}^{\policyoveroption{}}}
\newcommand{\optimalbehaviorQ}{Q^{\pi_{\omega}^*}}
\newcommand{\optimalskipQ}{\Tilde{Q}^{\pi_{\omega}^*}}
\newcommand{\skipQb}{\Tilde{Q}^{\policyoveroption{}}_{(b)}}
\newcommand{\behaviorQparam}{Q^{\policyoveroption{}}(s,a ; \theta)}
\newcommand{\skipQparam}{\Tilde{Q}^{\policyoveroption{}}(s,\omega_{aj};\phi)}

\usepackage[textsize=tiny]{todonotes}

\usepackage{newfloat}
\usepackage{listings}
\DeclareCaptionStyle{ruled}{labelfont=normalfont,labelsep=colon,strut=off} % DO NOT CHANGE THIS
\floatstyle{ruled}
\newfloat{listing}{tb}{lst}{}
\floatname{listing}{Listing}
\title{Learning Uncertainty-Aware Temporally-Extended Actions}
\author{
    Joongkyu Lee\textsuperscript{\rm 1}\equalcontrib,
    Seung Joon Park\textsuperscript{\rm 2}\equalcontrib,
    Yunhao Tang\textsuperscript{\rm 3},
    Min-hwan Oh\textsuperscript{\rm 1}
}
\affiliations{
    \textsuperscript{\rm 1}Seoul National University\\
    \textsuperscript{\rm 2}Samsung Research\\
    \textsuperscript{\rm 3}Google DeepMind\\
    jklee0717@snu.ac.kr,
    soonjun.park@samsung.com,
    robintyh@deepmind.com,
    minoh@snu.ac.kr 
}

\usepackage{bibentry}
\begin{document}

\maketitle

\begin{abstract}
In reinforcement learning, temporal abstraction in the action space, exemplified by action repetition, is a technique to facilitate policy learning through extended actions. However, a primary limitation in previous studies of action repetition is its potential to degrade performance, particularly when sub-optimal actions are repeated. 
This issue often negates the advantages of action repetition. 
To address this, we propose a novel algorithm named \textbf{U}ncertainty-aware \textbf{T}emporal \textbf{E}xtension (\Algname{}). \Algname{} employs ensemble methods to accurately measure uncertainty during action extension. This feature allows policies to strategically choose between emphasizing exploration or adopting an uncertainty-averse approach, tailored to their specific needs. We demonstrate the effectiveness of \Algname{} through experiments in Gridworld and Atari 2600 environments. Our findings show that \Algname{} outperforms existing action repetition algorithms, effectively mitigating their inherent limitations and significantly enhancing policy learning efficiency.
\end{abstract}

%%%%%%%%%%%%%%%%%%%%%%%%%%%%%%%%%%%%%%%%%%%%%%%%%%%%%%%%%%%%%%%%%%%%%%%%%%%%%%%%%%%%%%%%%%%
\section{Introduction}
Temporal abstraction is a promising approach to solving complex tasks in reinforcement learning (RL) with complex structures and long horizons~\cite{fikes1972learning, dayan1992feudal, parr1997reinforcement, sutton1999between, precup2000temporal, bacon2017option, barreto2019option, machado2021temporal}. 
Hierarchical reinforcement learning (HRL) enables the decomposition of this sequential decision-making problem into simpler lower-level actions or subtasks.
Intuitively, an agent explores the environment more effectively when operating at a higher level of abstraction and solving smaller subtasks~\cite{machado2021temporal}. 
One of the most prominent approaches for HRL is the \textit{option} framework~\cite{sutton1999between, precup2000temporal}, which describes the hierarchical structure in decision making in terms of temporally-extended courses of action.
Temporally-extended actions have been shown to speed up learning, potentially providing more effective exploration compared to single-step explorative action and requiring a smaller number of high-level decisions when solving a problem~\cite{stolle2002learning, biedenkapp2021temporl}. 
From a cognitive perspective, such observations are also coherent with how humans learn, generalize from experiences, and perform abstraction over tasks~\cite{xia2021temporal}. 

There has been a line of works that propose repetition of action for an extended period as a specialized form of temporal abstraction~\cite{lakshminarayanan2017dynamic, sharma2017learning, dabney2020temporally, metelli2020control, biedenkapp2021temporl, park2021time}.\footnote{
In fact, action repetition for a fixed number of steps was one of the strategies deployed in solving Atari~2600 games~\cite{mnih2015human, machado18arcade}. Despite its simplicity, the action repetition provided sufficient performance gains so that almost all modern methods of solving Atari games are still implementing such action repetitions.} 
Hence, the action-repetition methods address the problem of learning when to perform a new action while repeating an action for multiple time-steps~\cite{dabney2020temporally, biedenkapp2021temporl}.
The \rc{}, the interaction steps to repeat the same action, 
is learned by an agent along with what action to execute~\cite{sharma2017learning, biedenkapp2021temporl}. 
As shown by the improved empirical performances~\cite{dabney2020temporally, biedenkapp2021temporl}, these action repetition approaches can be well justified by the \textit{commitment} to action for deriving a deeper exploration. 
These approaches can help suppress the dithering behavior of the agent that can result in short-sighted exploration in a local neighborhood. 

However, simple action repetition alone cannot guarantee performance improvement. 
Repetition of a sub-optimal action for an extended period can lead to severe deterioration in the performance.
For example, a game may terminate due to reckless action repetition when an agent is in a dangerous region. A more uncertainty-averse behavior would be helpful in this scenario.
On the other hand, an agent may linger in the local neighborhood due to a lack of optimism, especially in sparse reward settings. In that case, a more exploration-favor behavior can be beneficial.
In either case, a suitable control of uncertainty of value estimates over longer horizons can be a crucial element.
In particular, the calibration of how much exploration the agent can take, or how uncertainty-averse the agent should be, can definitely depend on an environment.
Thus, the degree of uncertainty to be considered should be adaptive depending on the environment.
To this end, we propose to account for uncertainties when repeating actions. 
To our best knowledge, consideration of uncertainty in the future when instantiating action repetition has been not addressed previously.
Such consideration is essential in action repetition in both uncertainty-averse and exploration-favor environments.

In this paper, we propose a novel method that learns to repeat actions while incorporating the estimated uncertainty of the repeated action values. 
We can either impose aggressive or uncertainty-averse exploration by controlling the degree of uncertainty in order to take suitable uncertainty-aware strategy for the environment. 
Through extensive experiments and ablation studies, we demonstrate the efficacy of our proposed method and how it  enhances the performances of deep reinforcement learning agents in various environments.  
In comparison with the benchmarks, we show that our proposed method outperforms baselines, consistently outperforming the existing action repetition methods.
Our contributions are: 
\begin{itemize}
    \item We present a novel framework that allows the agent to repeat actions in a uncertainty-aware manner using an ensemble method. 
    Suitably controlling the amount of uncertainty induced by repeated actions, our proposed method learns to choose \rc{} and learns how \textit{optimistic} or \textit{pessimistic} it should be, hence enabling efficient exploration. 
    \item Our method yields a salient insight that it is beneficial to consider environment-inherent uncertainty preference.
    Some environments are uncertainty-favor (Chain MDP), and some are uncertainty-averse (Gridworlds).
    \item In a set of testing environments, we show \Algname{} consistently outperforms all of the existing action-repetition baselines, such as \DAR{}, \TEE{}, \DQN{}, \BootstrappedDQN{}, in terms of final evaluation scores, learning speed, and coverage of state-spaces.  
\end{itemize}
%%%%%%%%%%%%%%%%%%%%%%%%%%%%%%%%%%%%%%%%%%%%%%%%%%%%%%%%%%%%%%%%%%%%%%%%%%%%%%%%%%%%%%%%%%%

\section{Related Work}
\label{Related_Work}
\textbf{Temporal Abstraction and Action Repetition.}
Temporal abstractions can be viewed as an attempt to find a time scale that is adequate for describing the actions of an AI system~\cite{precup2000temporal}. 
The options framework~\cite{sutton1999between, precup2000temporal, bacon2017option} formalizes the idea of temporally-extended actions.
An MDP endowed with a set of options are called Semi-Markov Decision Process (SMDP) which we define in Preliminaries. The generalization of conventional action-value functions for the options framework is called \textit{option}-value functions~\cite{sutton1999between}. The mapping from states to probabilities of taking an option is called policy over options. In the options framework, the agent attempts to learn a policy over options that maximizes the option-value functions.

One simple form of an option is repeating a primitive action for certain number of steps~\cite{schoknecht2002speeding}. 
Action repetition has been widely explored in the literature~\cite{lakshminarayanan2017dynamic, sharma2017learning, dabney2020temporally, metelli2020control, biedenkapp2021temporl, park2021time}. Action repetition has been empirically shown to induce deeper exploration \cite{dabney2020temporally} and lead to efficient learning by reducing the granularity of control~\cite{lakshminarayanan2017dynamic, sharma2017learning,  metelli2020control, biedenkapp2021temporl}. 
Action repetition can be implemented by deciding the \rc{} of an action which is either sampled from a distribution~\cite{dabney2020temporally} or returned by a policy~\cite{lakshminarayanan2017dynamic, sharma2017learning}. 
The closest related to our work is~\citet{biedenkapp2021temporl}. 
They proposed an algorithm called \TempoRL{} that not only selects an action in a state but also for how long to commit to that action. 
\TempoRL{}~\cite{biedenkapp2021temporl} proposes a hierarchical structure in which \textit{behavior} policy determines the action $a$ to be played given the current state $s$, and a \textit{skip} policy determines how long to repeat this action.
However, our main intuition is that simply repeating the chosen action is not enough. 
We may encounter undesirable states while repeating the action.
This could lead to catastrophic failure when an agent enters a ``risky'' area (refer Gridworlds experiments). 
Our method has been shown to effectively manage this issue by quantifying the uncertainty of the option in form of repeating actions.

\textbf{Uncertainty in Reinforcement Learning.}
Recently, many works have made significant advances in empirical studies by quantifying and incorporating uncertainty~\cite{osband2016deep, bellemare2016unifying, badia2020agent57, lee2022offline}.
There are two types of uncertainty: aleatoric and epistemic.
Aleatoric uncertainty is the uncertainty caused by the uncontrollable stochastic nature of the environment and cannot be reduced.
Epistemic uncertainty is caused by the current imperfect training of the neural network and can be reducible.

One mainstream of estimating the uncertainty in deep RL relies on bootstrapping.
\citet{osband2016deep} introduced Bootstrapped DQN as a method for effcient exploration.
This approach is a variation of the classic DQN neural network architecture, which has a shared torso with $K \in \mathbb{Z}^{+}$ heads.
\citet{anschel2017averaged, peer2021ensemble} leveraged an ensemble of Q-functions to mitigate overestimation in DQN.
In this paper, we propose an algorithm that quantifies uncertainty of Q-value estimates of the states reached under the repeated-action.
This algorithm utilizes multiple randomly-initialized bootstrapped heads that stretch out from a shared network, providing multiple estimates of the \textit{option}-value function. 
The variance between these estimates is then used as a measure of uncertainty. 
Notably, this approach allows us to capture both aleatoric and epistemic uncertainty.
Then, we establish a UCB-style~\citep{auer2002finite, audibert2009exploration} option-selecting algorithm that simply adds the estimated uncertainty to the averaged ensemble Q-values and chooses an action that maximizes the quantity~\citep{chen2017ucb, peer2021ensemble}.
%%%%%%%%%%%%%%%%%%%%%%%%%%%%%%%%%%%%%%%%%%%%%%%%%%%%%%%%%%%%%%%%%%%%%%%%%%%%%%%%%%%%%%%%%%%
\section{Preliminaries and Notations}
\label{sec:preliminaries}
In reinforcement learning, an agent interacts with an environment whose underlying dynamics is modeled by a Markov Decision Process (MDP) \cite{puterman2014markov}. 
The tuple $ \langle  \mathcal{S}, \mathcal{A}, P, R, \gamma \rangle $ defines an MDP $\mathcal{M}$, where $\mathcal{S}$ is a state space, $\mathcal{A}$ is an action space, $P:\mathcal{S} \times \mathcal{A} \rightarrow \mathcal{S} $ is a transition dynamics function, $r: \mathcal{S} \times \mathcal{A} \rightarrow \mathbb{R}$ is a reward function, and $\gamma \in [0,1]$ is the discount factor.
We consider a Semi-Markov Decision Process (SMDP) model to incorporate the options framework~\cite{sutton1999between, precup2000temporal}.
An SMDP is an original MDP with a set of options, i.e., $ \mathcal{M}_o := \langle  \mathcal{S}, \Omega, P_o, R_o \rangle $, 
where $\omega \in \Omega$ is an option in the option space, $P_o(s'\mid s,\omega) : \mathcal{S} \times \Omega \rightarrow \mathcal{S}$ is the probability of transitioning from state $s$ to state $s'$ after taking an option $\omega$ and $R_o: \mathcal{S} \times \Omega \rightarrow \mathbb{R}$ is the reward function for the option.

For any set $\mathcal{X}$, let $\mathcal{P}(\mathcal{X})$ denote the space of probability distributions over $\mathcal{X}$.
Then a policy over option $\policyoveroption{}: \mathcal{S} \rightarrow \mathcal{P}(\Omega) $ assigns a
probability to an option conditioned on a given state. 
Our goal is to learn a policy $\policyoveroption{}$ that maximizes the expectation of discounted return starting from a initial state $s_0$; then, define the value functions $V^{\policyoveroption{}}(s_0) = \mathbb{E}_{\policyoveroption{}} [\sum_{t=0}^{\infty}\gamma^t R_t \mid s_0]$, the action-value functions $\behaviorQ{}(s_0, a) = \mathbb{E}_{\policyoveroption{}} [\sum_{t=0}^{\infty}\gamma^t R_t \mid s_0, a]$, or the option-value functions $\skipQ{}(s_0, \omega) = \mathbb{E}_{\policyoveroption{}} [\sum_{t=0}^{\infty}\gamma^t R_t \mid s_0, \omega]$.

In general, options depend on the entire \textit{history} between time step $t$ when they were initiated and the current time step $t+k$, $h_{t:t+k} := s_ta_ts_{t+1}...a_{t+k-1}s_{t+k}$.
Let $\mathcal{H}$ be the space of all possible histories $h$, then a \textit{semi-Markov option} $\omega$ is a tuple $\omega := \langle \mathcal{I}_{o}, \pi_{o}, \beta_{o} \rangle$, where $ \mathcal{I}_{o} \subset \mathcal{S}$ is an initiation set, $\pi_{o} : \mathcal{H} \rightarrow \mathcal{P}(\mathcal{A})$ is an \textit{intra-option} policy, and $\beta_{o}: \mathcal{H} \rightarrow [0,1]$ is a termination function. 
In this framework, we define an action repeating option to be $\omega_{aj} := \langle \mathcal{S}, \bm{1}_a, \beta(h)=\bm{1}_{|h|=j} \rangle$, in which $h \in \mathcal{H}$ and $\bm{1}_a$ indicates $|\mathcal{A}|$-dimensional vector where the element corresponding to $a$ is 1 and 0 otherwise.
This action repeating option takes action $a$ for $j$ times and then terminates.

When an agent plays a chosen action for \rc{} $j$, total of $\frac{j\dot(j+1)}{2}$ skip-transitions are observed and stored in the replay buffer~\cite{biedenkapp2021temporl}.
Specifically, when repeating the action for $j$ times from state $s$, we can also experience ($s \rightarrow s'_{(1)}), (s \rightarrow s'_{(2)}), \dots, (s'_{(1)} \rightarrow s'_{(2)}), \dots, (s'_{(j-1)} \rightarrow s'_{(j)})$, in total $\frac{j\cdot(j+1)}{2}$ transitions.
We leverage these transitions to update option-values. 
Consequently, the observations for short extensions are updated more frequently, leading to smaller uncertainties for short extensions and larger uncertainties for long extensions.
\begin{algorithm}[t!]
    \caption{UTE: Uncertainty-aware Temporal Extension}\label{brief_algo}
    \begin{algorithmic}[1]
    \State \textbf{Input:} uncertainty parameter $\lambda$, the number of output heads of option-value functions $B$.
    \State \textbf{Initialize:} $\behaviorQ{}$, $\{\skipQb\}_{b=1}^B$.
    \For{episode = $1, \dots ,K$}
        \State {Obtain initial state $s$ from environment}
        \Repeat
            \State { $a \leftarrow \epsilon$-greedy $\arg\!\max_{a'}\,\behaviorQ{} (s,a)$} 
            \State {Calculate $\hat{\mu}_{\policyoveroption{}}(s,\omega_{aj})$, $\hat{\sigma}^{2}_{\policyoveroption{}}(s,\omega_{aj})$ by Eq.~\eqref{eq:mu_sigma}.}
            \State { $j \leftarrow \arg\!\max_{j'} \,\{\hat{\mu}_{\policyoveroption{}}(s,\omega_{aj'}) + \lambda \hat{\sigma}_{\policyoveroption{}}(s,\omega_{aj'})\}$ }
            \While {$j \neq 0$ and $s$ is not terminal}
                \State {Take action $a$ and observe $s'$, $r$}
                \State $s \leftarrow s'$, $j \leftarrow j-1$
            \EndWhile
        \Until {episode ends}
    \EndFor
    \end{algorithmic}
\end{algorithm}
%
%%%%%%%%%%%%%%%%%%%%%%%%%%%%%%%%%%%%%%%%%%%%%%%%%%%%%%%%%%%%%%%%%%%%%%%%%%%%%%%%%%%%%%%%%%%
\section{Uncertainty-aware Temporal Extension}
\label{Main Algorithm}
In this section, we propose our algorithm \Algname: \textbf{U}ncertainty-aware \textbf{T}emporal \textbf{E}xtension, which repeats the action in consideration of uncertainty in Q-values.
We first demonstrate temporally-extended \Q{}-learning by decomposing the action repeating option.
We then describe how we estimate the uncertainty of an option-value function $\skipQ{}$ by utilizing the ensemble method to select an \rc{} $j$ in consideration of uncertainty. 
We additionally show that \textit{n}-step targets can be used for learning the action-value function $\behaviorQ{}$ without worrying about off-policy correction.

\subsection{Temporally-extended \Q{}-Learning} \label{Temporally}
In this work, we mainly depend on techniques based on the \Q{}-learning algorithm \cite{watkins1992q}, which seeks to approximate the Bellman optimality operator to learn the optimal policy:
\begin{definition}
We define the optimal action-value function $\optimalbehaviorQ{}$ and the optimal option-value function $\optimalskipQ{}$ respectively as
\begin{align}
    \optimalbehaviorQ{}\!(s,a)\!\!=& \mathbb{E}_{s'_{(1)} \sim P}\!\!\left[R(s,a)\!+\!\gamma \underset{a'}{\max}\,\optimalbehaviorQ{}(s'_{(1)},a')  \right],  \label{eq:behavior_Qlearning_TempoRL}\\
    \optimalskipQ{}\!(s,\omega_{aj})\!\!=& \mathbb{E}_{s'_{(j)} \sim P_o }\!\!\left[ R_o(s,\omega_{aj})\!+\!\gamma^{j} \underset{\omega'}{\max}\,\optimalskipQ{}(s'_{(j)},\omega') \right], \label{eq:skip_Qlearning_TempoRL}
\end{align}
\end{definition}
where $s'_{(0)}$ and $s'_{(j)}$, respectively, indicate one-step and $j$-step later state from the state $s$    .
In practice, it is common to use a function approximator to estimate each \Q{}-value, $\behaviorQparam{} \approx \optimalbehaviorQ{}(s,a)$ and $\skipQparam{} \approx \optimalskipQ{}(s,\omega_{aj})$.
We use two different neural network function approximators parameterized by $\theta$ and $\phi$ respectively.

\textbf{Option Decomposition.} Learning the optimal policy over options, instead of the optimal action policy, has the same effect as enlarging the action space from $|\mathcal{A}|$ to $|\mathcal{A}| \times |\mathcal{J}|$, where $\mathcal{J}=$\{1, 2, ... , max repetition\}.
Generally, inaccuracies in \Q{}-function estimations can cause the learning process to converge to a sub-optimal policy, and this phenomenon is amplified in situations with large action spaces~\cite{thrun1993issues, zahavy2018learn}.
Therefore, we consider decomposed policy over option~\cite{biedenkapp2021temporl}, $\policyoveroption{}(\omega_{aj}\mid s) := \actionpolicy{}(a\mid s) \cdot \extensionpolicy{}(j\mid s,a)$, in which an action policy $\actionpolicy{}(a\mid s):\mathcal{S \rightarrow \mathcal{P}(A)}$ assigns some probability to each action conditioned on a given state, and then an extension policy $\extensionpolicy{}(j\mid s,a):\mathcal{S \times A} \rightarrow \mathcal{P}(\mathcal{J})$ assigns some probability to each \rc{} conditioned on a given state and action.
Note that there exists a hierarchy between decomposed policies $\actionpolicy{}$ and $\extensionpolicy{}$, thus, $\actionpolicy{}$ always has to be queried before $\extensionpolicy{}$ at every time an option initiates.
The agent first chooses an action $a$ from action policy $\actionpolicy{}$ based on the action-value function $\behaviorQ{}$ (e.g. $\epsilon$-greedy). 
Then, given this action $a$, it selects \rc{} $j$ from $\extensionpolicy{}$ according to the option-value function $\skipQ{}$.

By decomposing the policy over option $\policyoveroption{}$, we can decrease the search space from $|\mathcal{A}| \times |\mathcal{J}|$ to $|\mathcal{A}| + |\mathcal{J}|$. 
We empirically show that decomposing option can stabilize the \Q{}-learning in Appendix.
However, this learning process may converge to a sub-optimal policy because it is intractable to search all the possible combinations of actions and \rc{}s $(a,j)$.
The agent may repeat the sub-optimal action excessively or sometimes be overly myopic.
Our algorithm can mitigate this issue by controlling the level of uncertainty when executing the extension policy $\pi_e$.
\begin{proposition}\label{prop:optimal_value} 
In a Semi-Markov Decision Process (SMDP), let an option $\omega \in \Omega$ be the action repeating option defined by action $a$ and \rc{} $j$, i.e. $\omega_{aj} := \langle \mathcal{S}, \bm{1}_a, \beta(h)=\bm{1}_{h=j} \rangle $. 
For all $\omega \in \Omega$, a policy over option, $\policyoveroption$, can be decomposed by an action policy $\actionpolicy{}(a \mid s):\mathcal{S \rightarrow \mathcal{P}(A)}$ and an extension policy $\extensionpolicy{}(j \mid s,a):\mathcal{S \times A} \rightarrow \mathcal{P}(|J|)$, i.e. $\policyoveroption(\omega_{aj} \mid s) := \actionpolicy{}(a \mid s) \cdot \extensionpolicy{}(j \mid s,a)$.
Then, for the corresponding optimal policy $\policyoveroption^*$, the following holds:
\begin{align*}
        V^{\pi_{\omega}^*}(s) = \max_{\omega_{aj}}\,\optimalbehaviorQ{}(s,\omega_{aj}) = \max_{a}\,\optimalbehaviorQ{}(s,a).
\end{align*}
\end{proposition}
Proposition 1 implies that the target value for the option selection of repeated actions can be the same as the target for a single-step action selection within the option.
In our implementation, we use $\underset{a'}{\max}\,\optimalbehaviorQ{}(s'_{(j)},a')$ instead of $\underset{\omega'}{\max}\,\optimalskipQ{}(s'_{(j)},\omega'_{aj})$ for the target value in Eq.\eqref{eq:skip_Qlearning_TempoRL}.
This can stabilize the learning process by sharing the same target.

\subsection{Ensemble-based Uncertainty Quantification}
\label{Risk}
In the previous action repetition methods~\cite{lakshminarayanan2017dynamic, sharma2017learning, dabney2020temporally, biedenkapp2021temporl}, they extend the chosen action without considering uncertainty which could easily run to failure. 
The only situation where these problems do not occur is when their \repetition{} policies are optimal, which means they need to expect the $j$ step later state precisely.
However, it is improbable in the sense that this situation rarely occurs in the learning process.
In order to solve this problem, we propose a strategy of choosing a \rc{} $j$ in an uncertainty-aware manner. 
\Algname{} is a uncertainty-aware version of the \TempoRL{} \cite{biedenkapp2021temporl}.
Our main intuition is that it is crucial to consider the uncertainty of option-value functions $\skipQ{}$, when selecting \rc{} $j$ by \repetition{} policy~$\extensionpolicy{}$. 

We use the ensemble method, which has recently become prevalent in RL~\cite{osband2016deep, da2020uncertainty, bai2021principled}, to estimate uncertainty in our estimated option-value functions.
We use a network consisting of a shared architecture with $B$ independent. ``head'' branching off from the shared network.
Each head corresponds to a option-value function, $\skipQb{}$, for $b\in\{1,2,\dots,B\}$. 
Each head is randomly-initialized and trained by different samples from an experience buffer. 
Unlike Bootstrapped DQN (\BootstrappedDQN{})~\cite{osband2016deep} where each one of the value function heads is trained against its own target network, our \Algname{} trains each value function head against the same target.
If each head has its own target head respectively, since the objective function of neural networks is generally non-convex, each Q-value may converge to different modes.
In this case, as training the policy, the estimated uncertainty of option Q-value, $\hat{\sigma}_{\policyoveroption{}}$, could not converge to zero. This means that it is unable to learn an optimal policy.
Therefore, using the same target is one of the key points of our implementation. 

Given state $s$ and action $a$, $\skipQb{}$-values are aggregated by \rc{}~$j$ to estimate mean and variance as follows:
\begin{align}
    \hat{\mu}_{\policyoveroption{}}(s,\omega_{aj}) &:= \dfrac{1}{B}\sum_{b=1}^{B}\skipQb{}(s,\omega_{aj})  \label{eq:mu_hat}\\
    \hat{\sigma}^{2}_{\policyoveroption{}}(s,\omega_{aj}) &:= \dfrac{1}{B}\sum_{b=1}^{B}(\skipQb{}(s,\omega_{aj}) )^{2} - (\hat{\mu}_{\policyoveroption{}}(s,\omega_{aj}))^{2}  \label{eq:mu_sigma}
\end{align}
Then, we define \textit{uncertainty-aware}~\repetition{} policy~$\extensionpolicy{}$, which takes \rc{}~$j$ deterministically given state and action, by introducing the~\riskparameter{}~$\lambda\in\mathbb{R}$:
\begin{align}
j = \underset{j'\in \mathcal{J}}{\arg\!\max}\,\{\hat{\mu}_{\policyoveroption{}}(s,\omega_{aj'}) + \lambda \hat{\sigma}_{\policyoveroption{}}(s,\omega_{aj'})\}. \nonumber 
\end{align}
where $\lambda$ indicates the level of uncertainty to be considered. 
The positive $\lambda$ induces more aggressive exploration, and the negative one causes uncertainty-averse exploration.
%%%%% N step
\subsection{Multi-step Q-Learning}
\label{Nstep}
We make use of $n$-step \Q-learning \cite{sutton1988learning} to learn both $\behaviorQ{}$ and $\skipQ{}$, whereas \TempoRL{} \cite{biedenkapp2021temporl} used it only for updating $\skipQ{}$. 
We found that \textit{n}-step targets can also be used to update $\behaviorQ{}$-values without any off-policy correction~\cite{harutyunyan2016q}, e.g., importance sampling. 
Given the sampled $n$-step transition $\tau_t=(s_t,a_t,R_o(s_t,o_{an}),s_{t+n})$ from replay buffer $\mathcal{R}$, as long as $n$ is smaller than or equal to the current \repetition{} policy $\extensionpolicy{}$'s output $j$, the transition $\tau_t$ trivially follows our target policy $\policyoveroption{}$. 
Thus, $\tau_t$ can be directly used to update the action-value function $\behaviorQ{}$. 
Instead of one step \Q{}-learning in Eq.\eqref{eq:behavior_Qlearning_TempoRL}, 
\Algname{} uses $n$-step \Q-Learning to update $\behaviorQ{}$:
\begin{align*} 
 \mathcal{L}_{\behaviorQ{}}(\theta) &= \mathbb{E}_{\tau_{t} \sim \mathcal{R}} \Big[(\Q^{\pi_{\omega}}(s_t, a_t ; \theta) - \sum_{k=0}^{n-1}\gamma^k r_{t+k} \\
 &- \gamma^n \max_{a'}\,\optimalbehaviorQ{}(s_{t+n},a'; \Bar{\theta})  )^2 \,\Bigl\vert\, n \leq j \sim \extensionpolicy{} \Big] 
\end{align*}
where $\bar{\theta}$ are the delayed parameters of action-value function $\behaviorQ{}$ and $j \sim \extensionpolicy{}(j_t\mid s_t, a_t)$.
In general, $n$-step returns can be used to propagate rewards faster~\cite{watkins1989learning, peng1994incremental}.
It mitigates the overestimation problem in Q-learning as well~\cite{meng2021effect}.
We empirically illustrate that $n$-step learning leads to faster learning in Figure~\ref{fig:ablation_nstep}.
Note that we don't need to pre-define $n$ because it is dynamically determined by current \repetition{} policy $\extensionpolicy{}$. 

\subsection{Adaptive Uncertainty Parameter}
Instead of fixing the uncertainty parameter $\lambda$ during the learning process, we propose the adaptive selection of $\lambda$ utilizing a non-stationary multi-arm bandit algorithm, as described in~\citep{badia2020agent57}. 
Consider $\Lambda$ as the predefined set of uncertainty parameters. At the onset of each episode $k$, the bandit selects an arm, denoted by $\lambda_k \in \Lambda$, and subsequently receives feedback in the form of episode returns $R_k(\lambda_k)$. 
Given that the reward signal $R_k(\lambda_k)$ is non-stationary, we employ a sliding-window UCB combined with $\epsilon_{ucb}$-greedy exploration to optimize the process. 
Further details regarding the algorithms can be found in Appendix.

%%%%%%%%%%%%%%%%%%%%%%%%%%%%%%%%%%%%%%%%%%%%%%%%%%%%%%%%%%%%%%%%%%%%%%%%%%%%%%%%%%%%%%%%%%%

\section{Experiments} \label{sec5:experiments}
In this section, we present three principal experimental results: Chain MDP, Gridworlds, and Atari 2600 games, as described in Machado et al. (2018)~\cite{machado18arcade}.
Initially, we confirm our hypothesis that a positive $\lambda$ foster more aggressive exploration (ChainMDP), while a negative one results in uncertainty-averse exploration (Gridworlds).
Subsequently, we demonstrate the significant impact of a well-tuned $\lambda$ on performance in more complex environments and illustrate that the adaptive selection of $\lambda$ consistently outperforms other baseline measures (Atari 2600 games).
To ensure a fair comparison, we explored a considerable range of hyperparameters to identify the most optimal value for each algorithm (Refer Table~\ref{tab:nchain_auc_all}, \ref{tab:grid_tee_mu}, \ref{tab:atari_tee_mu}, and~\ref{tab:atari_result_rs} in Appendix)

\begin{figure}[ht]
    \centering
    \includegraphics[width=0.8\linewidth]{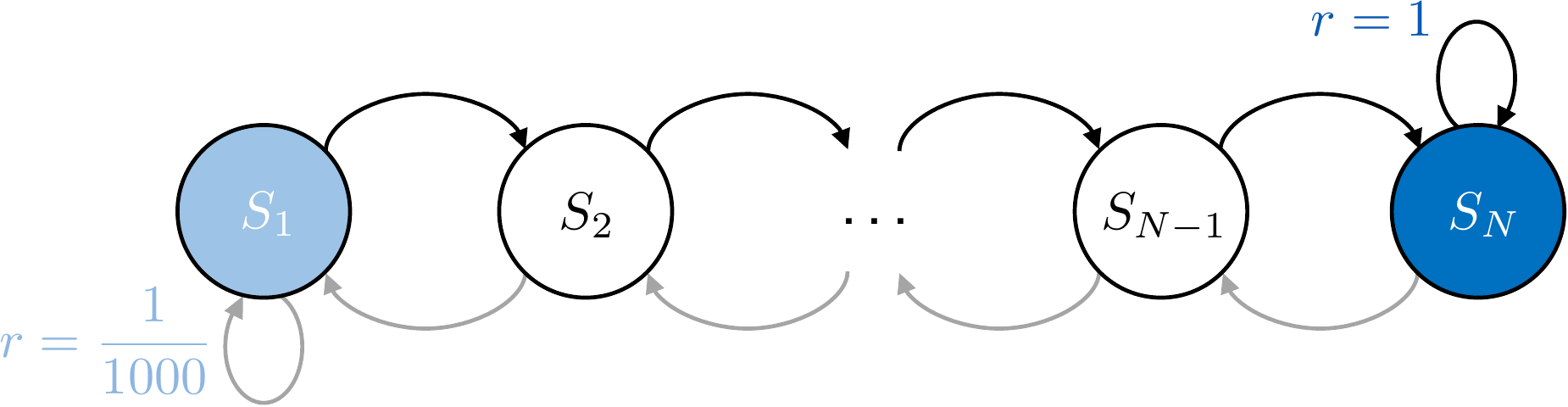}
    \captionof{figure}{Chain MDP}
    \label{fig:chainMDP}
\end{figure}
\subsection{Chain MDP} \label{exp:chainMDP}
We experimented in the Chain MDP environment as described in Figure~\ref{fig:chainMDP}~\cite{osband2016deep}. 
There are two possible actions \{left, right\}.
If the agent reaches the left end ($s_1$) of the chain and performs a left action, a deceptive small reward ($0.001)$ is given.
And if the agent reaches the right end ($s_n$) of the chain and performs a right action, large reward ($1.0$) is given.
Thus, the optimal policy is to take only right actions.
Since the reward is very sparse, we need a ``deep'' exploration strategy to learn the optimal policy.
In this toy environment, we will verify our intuition that positive \riskparameter{} $\lambda$ induces deep exploration and as a result, show a better performance than other baselines, \DDQN{}~\cite{van2016deep}, \TEE{}~\cite{dabney2020temporally} and \TempoRL~\cite{biedenkapp2021temporl}. 

\textbf{Setup.} The agent interacts with the environment with a fixed horizon length, $N+8$, where $N$ is the chain length. 
Thus, the agent can obtain rewards from zero to 10 in each episode. 
We limited the maximum \rc{} as 10 for \TempoRL{} and \Algname{}.
\begin{table}[ht]
    \centering
    % \resizebox{\linewidth}{!}{
        \begin{tabular}{l|rrrr}
        \toprule
            Chain Length & 10 & 30 & 50 & 70  \\
        \midrule
        \TEE{}   & 0.654  & 0.427 & 0.434 & 0.131 \\
        \TempoRL  & 0.904    &  0.740  & 0.246 & 0.052  \\
        \Algname{} (ours)   & \textbf{0.919}  &  \textbf{0.758}  & \textbf{0.560} & \textbf{0.191}  \\
        \bottomrule
        \end{tabular}
        % }%
        \captionof{table}{
        Normalized AUC on Chain MDP over 20 runs
        }
        \label{tab:nchain_auc}
\end{table}

\textbf{Exploration-Favor.} Table~\ref{tab:nchain_auc} summarizes the results on various levels of chain length in terms of normalized area under the reward curve (AUC), comparing \Algname{} with the best \riskparameter{} (+2.0, the most optimistic $\lambda$) to \TEE{} and \TempoRL{}.
A reward AUC value closer to 1.0 indicates that the agent was able to find the optimal policy faster.
The total training episodes for calculating AUC was set to 1,000 across all chain lengths.
The results in the table show that \Algname{} outperforms the other two baselines notably throughout different chain lengths.
This implies that \Algname{} has a better exploration strategy which leads to higher reward even in the difficult settings (longer chain length).
We also point out that \Algname{} has a smaller variance than \TempoRL{} after it has reached the optimal reward of 10. 
This is mainly because \Algname{} can collect more diverse samples by exploratory extension policy $\extensionpolicy$, which may lead to better generalization and more accurate approximation to the optimal option-value function.

More importantly, we can encode the exploration-favor strategy by adjusting the \riskparameter{}, $\lambda$.
Note that we don't use $\epsilon$-greedy for the extension policy $\extensionpolicy$, whereas \TempoRL{} do.
In Appendix, Table~\ref{tab:nchain_auc_all} shows that more positive $\lambda$ achieves higher AUC scores.
When the agent selects a random action by the $\epsilon$-greedy action policy, it can explore deeper by being more optimistic, which leads to faster convergence to the optimal solution. 
An aggressive exploration strategy is beneficial because the environment has no risky area where the game terminates while repeating the action.

%%%%%%%%%%%%%%%%%%%%%%%%%%%%%%%%%%%%%%%%%%%%%%%%%%%%%%%%%%%%%%%%%%%%%%%%%%%%%%%%%%%%%%%%%%%

\begin{figure}[ht]
     \centering
     \begin{subfigure}[b]{0.18\textwidth}
         \includegraphics[width=\textwidth]{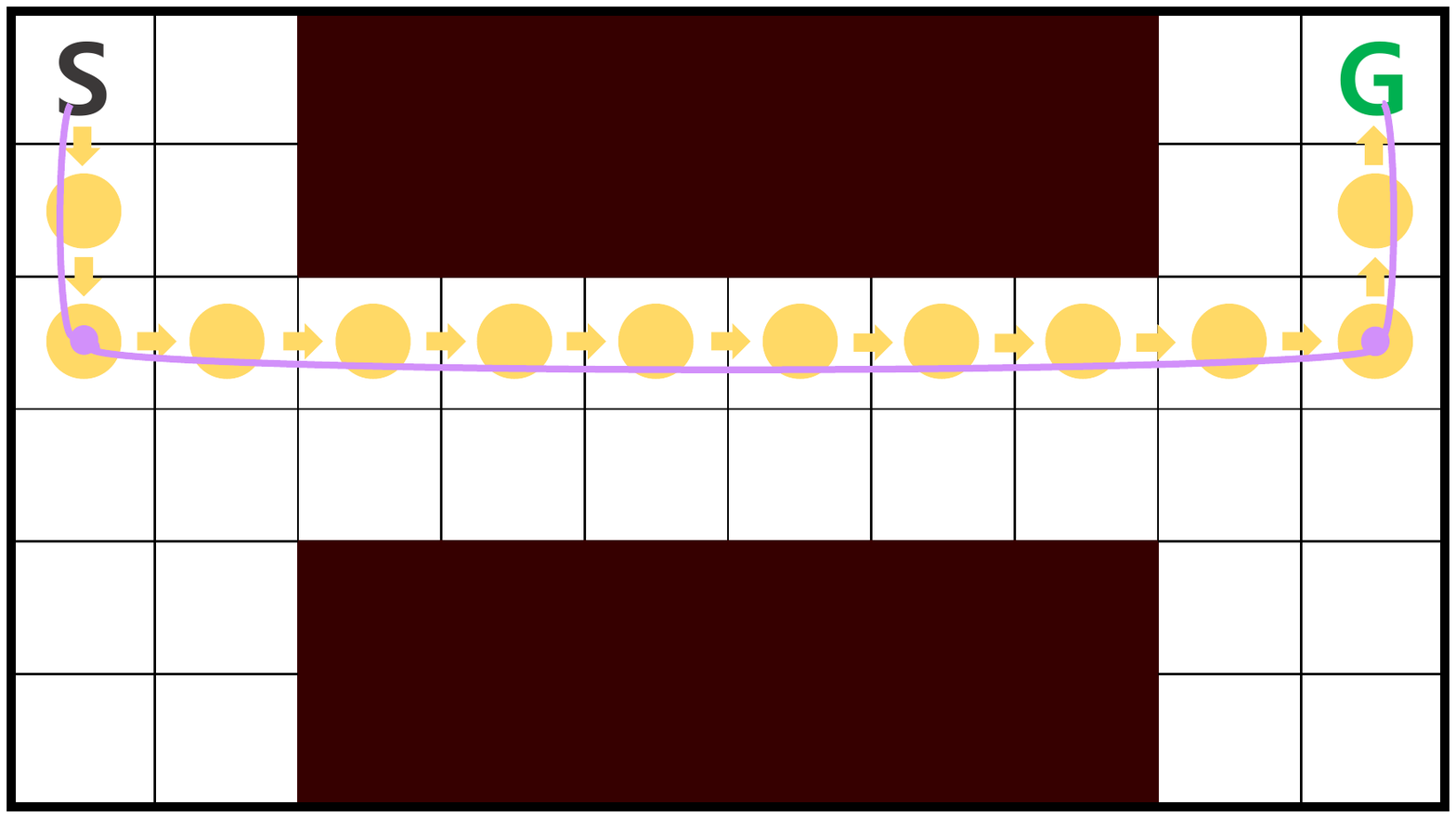}
         \caption{Bridge}
     \end{subfigure}
     \begin{subfigure}[b]{0.18\textwidth}
         \includegraphics[width=\textwidth]{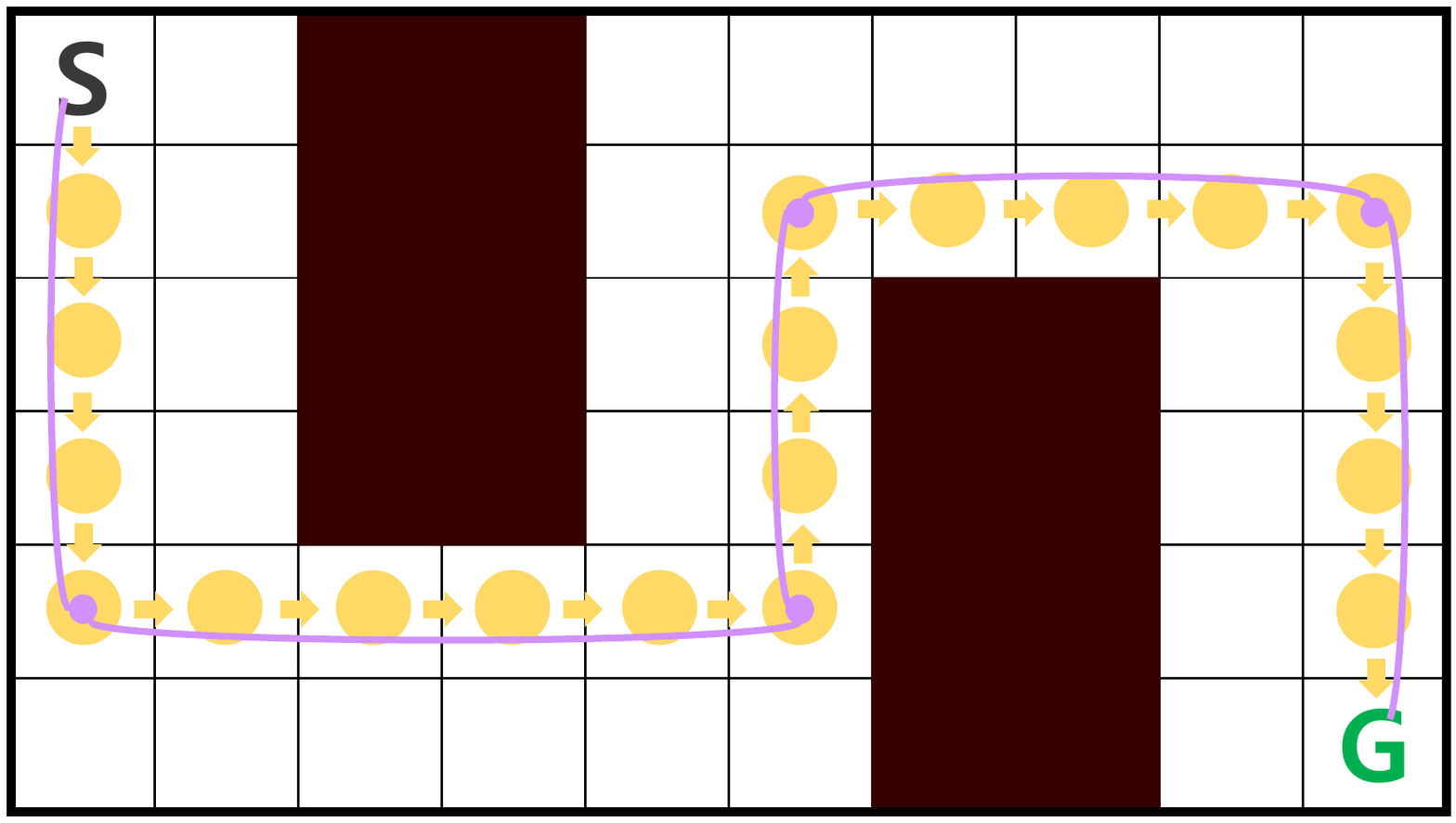}
         \caption{ZigZag}
     \end{subfigure}
    \caption{$6\times10$ Gridworlds. Agents have to reach a \textcolor{green}{goal} state~(\textcolor{green}{G}) from a \textcolor{gray}{starting} state (\textcolor{gray}{S}) detouring the lava. 
    Dots represent decision steps \textcolor{pink}{with} and \textcolor{brown}{without}  temporally-extended actions.} 
    \label{fig:three grid}
\end{figure}

\subsection{Gridworlds} \label{exp:gridworlds}
In this section, we analyze the empirical behavior of the various algorithms in the Gridworlds environment \textit{Lava} (Figure~\ref{fig:three grid}). 
It is a $6 \times 10$ grid with discrete states and actions. 
An agent starts in the top-left corner and must reach the goal to receive a positive reward (+1) while avoiding stepping into the lava (-1 reward) on its way.
In contrast to the chain MDP environment, since we have a risky area ``lava'', an uncertainty-averse strategy must be preferred.
We compare our method against vanilla~\DDQN~\cite{van2016deep} \TEE{}~\cite{dabney2020temporally} and~\TempoRL~\cite{biedenkapp2021temporl}. 

\textbf{Setup.} We trained all agents for a total of $3.0 \times 10^3$ episodes using 3 different types of $\epsilon$-greedy exploration  schedule: linearly decaying from 1.0 to 0.0 over all episodes, logarithmically decaying, and fixed $\epsilon=0.1$. 
We limited the maximum \rc{} to be 7. 
We use neural networks to learn \Q{}-value functions instead of tabular \Q{}-learning.

\begin{table}[ht!]
    \begin{center}\fontsize{9}{11}\selectfont
        \resizebox{\linewidth}{!}{
        \begin{tabular}{l|l|ccccc}
            \toprule
        Env & {$\epsilon$ decay} & \DDQN & \TempoRL & \TEE & \Algname \\
        \midrule
        \multicolumn{1}{l|}{\multirow{3}{*}{Bridge}} & Linear                          & 0.61                  & 0.44                     & 0.76 & \textbf{0.86} \\
        \multicolumn{1}{l|}{}                        & Log                             & 0.54                  & 0.32                     & 0.92 & \textbf{0.92} \\
        \multicolumn{1}{l|}{}                        & Fixed                           & 0.57                  & 0.41                     & 0.59 &  \textbf{0.83} \\ 
        \midrule
        \multicolumn{1}{l|}{\multirow{3}{*}{Zigzag}} & Linear                           & 0.38                  & 0.14                    & 0.62 &  \textbf{0.84} \\
        \multicolumn{1}{l|}{}                        & Log                            & 0.46                  & 0.12                      & 0.76 &  \textbf{0.89} \\
        \multicolumn{1}{l|}{}                        & Fixed                          & 0.34                  & 0.19                      & 0.36 &  \textbf{0.76}\\
        \bottomrule
        \end{tabular}
        }
        \vspace{5pt}
        \captionof{table}{
        Normalized AUC for reward across different $\epsilon$ exploration schedules over 20 random seeds.
        }
        \label{tab:gid_result}
    \end{center}
\end{table}
\begin{figure}[ht!]
    \centering
    \includegraphics[clip, trim=1.0cm 0.0cm 1.5cm 0.5cm,  width=0.95\linewidth]{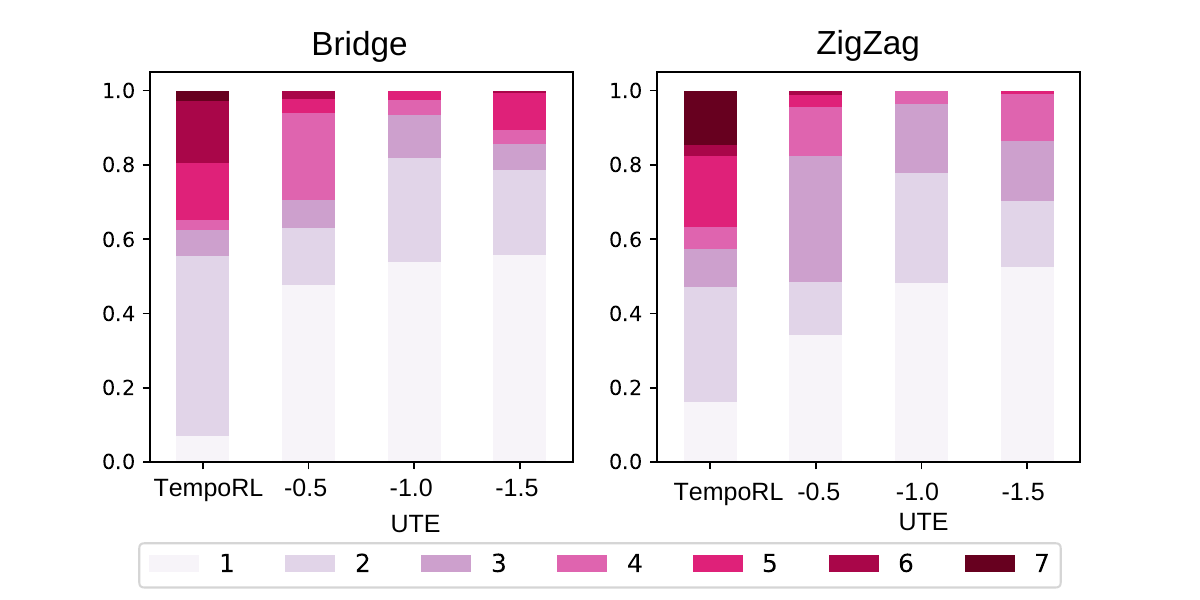}
    \captionof{figure}{Distributions of~\rc{} in Gridworlds.}
    \label{fig:lava1_hist.pdf}
\end{figure}

\textbf{Uncertainty-Averse.} We compare our \Algname{} to the other baselines in terms of normalized area under the reward curve for three different $\epsilon$-greedy schedules (see Table~\ref{tab:gid_result}). 
Across all $\epsilon$ exploration strategies, \Algname{} outperforms other methods while showing better performance as \riskparameter{} $\lambda$ becomes smaller (refer Table~\ref{tab:grid_tee_mu} in Appendix). 
This result supports our argument that a pessimistic strategy is preferred in environments with unsafe regions.
Furthermore, though exploration rate for $\actionpolicy$ is relatively large (e.g. fixed to $\epsilon = 0.1$), \Algname{} consistently shows good performance than others.

\begin{figure}[ht]
    \centering
    \includegraphics[clip, trim=0.0cm 10.9cm 16.0cm 2.1cm, width=0.9\linewidth]{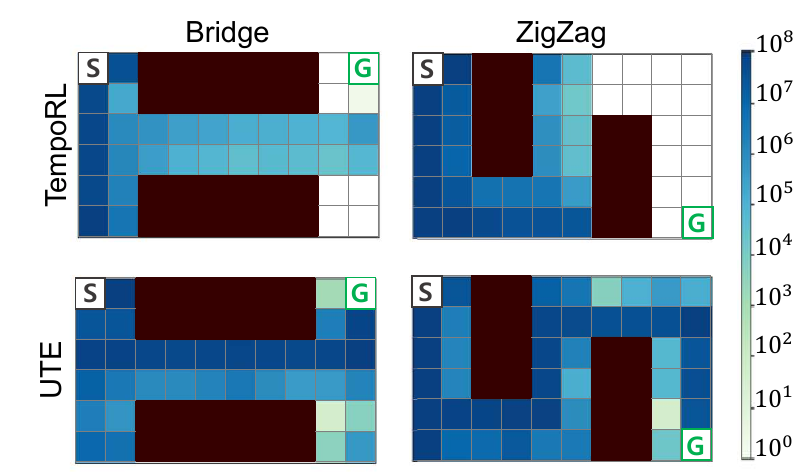}
    \caption{Coverage plots (right) on ZigZag environments. The blue represents states visited more often and white represents states rarely or never seen. See Appendix for the expanded version of the figures.}
    \label{fig:lava2,3_coverage}
\end{figure}

%%%%%%%%%%%%%%%%%%%%%%%%%%%%%%%%%%%%%%%%%%%%%%%%%%%%%%%
\begin{figure*}[ht]
    \centering
     \includegraphics[clip, trim=4.7cm 0.0cm 4.7cm 0.2cm, width=1.00\textwidth]{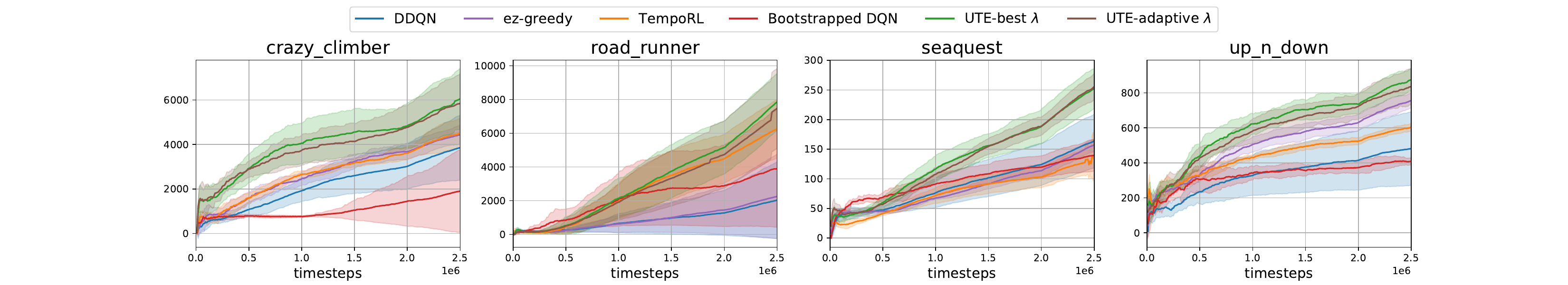}
     \caption{Learning curves of \Algname{} with best $\lambda$, \Algname{} with adaptive $\lambda$ and other baseline algorithms on Atari environments. The shaded area represents the standard deviation over 7 random seeds.}
     \label{fig:atari_best4_learning_curve} 
\end{figure*}
\begin{table*}[ht]
\centering 
\resizebox{\linewidth}{!}{
\begin{tabular}{l | rrrrrr | rrr} 
\toprule
\multirow{2}{*}{Environment} & \multirow{2}{*}{\DDQN} & \multirow{2}{*}{Fixed-$j$} & \multirow{2}{*}{\TEE{}} & \multirow{2}{*}{\DAR} & \multirow{2}{*}{\TempoRL} & \multirow{2}{*}{\BootstrappedDQN} & \multicolumn{3}{c}{\Algname{}}  \\
&&&&&&&$1$-step&  $n$-step & Adaptive $\lambda$\\
\midrule
 Crazy Climber  & 
            \setlength\extrarowheight{-3pt} \begin{tabular}[r]{@{}r@{}}5265.8\\{\tiny $\pm$ 4063.4} \end{tabular} & 
            \setlength\extrarowheight{-3pt}\begin{tabular}[r]{@{}r@{}}3731.1\\ {\tiny $\pm$ 2997.2}\end{tabular} & 
            \setlength\extrarowheight{-3pt}\begin{tabular}[r]{@{}r@{}}5295.1\\ {\tiny $\pm$3609.7}\end{tabular} & 
            \setlength\extrarowheight{-3pt}\begin{tabular}[r]{@{}r@{}}2059.1\\ {\tiny $\pm$1225.1}\end{tabular}  & 
            \setlength\extrarowheight{-3pt}\begin{tabular}[r]{@{}r@{}}4885.5\\ {\tiny $\pm$3378.3}\end{tabular}  & 
            \setlength\extrarowheight{-3pt}\begin{tabular}[r]{@{}r@{}}2961.6\\ {\tiny $\pm$3080.0}\end{tabular} & 
            \setlength\extrarowheight{-3pt}\begin{tabular}[r]{@{}r@{}}6761.9\\ {\tiny $\pm$5061.9}\end{tabular} & 
            \setlength\extrarowheight{-3pt}\begin{tabular}[r]{@{}r@{}}\textbf{8175.6}\\ {\tiny $\pm$5790.4}\end{tabular} &
            \setlength\extrarowheight{-3pt}\begin{tabular}[r]{@{}r@{}}7046.3\\ {\tiny $\pm$5350.0}\end{tabular} \\

 Road Runner  & 
            \setlength\extrarowheight{-3pt} \begin{tabular}[r]{@{}r@{}}3277.0\\{\tiny $\pm$4470.3} \end{tabular} & 
            \setlength\extrarowheight{-3pt} \begin{tabular}[r]{@{}r@{}}1230.3\\ {\tiny $\pm$1640.9}\end{tabular}  & 
            \setlength\extrarowheight{-3pt} \begin{tabular}[r]{@{}r@{}}3733.8\\ {\tiny $\pm$4716.5}\end{tabular}  & 
            \setlength\extrarowheight{-3pt} \begin{tabular}[r]{@{}r@{}}845.5\\ {\tiny $\pm$791.9}\end{tabular} & 
            \setlength\extrarowheight{-3pt} \begin{tabular}[r]{@{}r@{}}8131.5\\ {\tiny $\pm$4099.3}\end{tabular}  &
            \setlength\extrarowheight{-3pt} \begin{tabular}[r]{@{}r@{}}4976.8 \\ {\tiny $\pm$6032.6}\end{tabular}& 
            
            \setlength\extrarowheight{-3pt} \begin{tabular}[r]{@{}r@{}}4935.6\\ {\tiny $\pm$5206.4}\end{tabular} & 
            \setlength\extrarowheight{-3pt} \begin{tabular}[r]{@{}r@{}}\textbf{12323.2}\\ {\tiny $\pm$4177.1}\end{tabular}  &
            \setlength\extrarowheight{-3pt} \begin{tabular}[r]{@{}r@{}}10353.3\\ {\tiny $\pm$3283.3}\end{tabular}\\
            
 Sea Quest  & 
            \setlength\extrarowheight{-3pt} \begin{tabular}[r]{@{}r@{}}207.6\\{\tiny $\pm$124.5} \end{tabular} & 
            \setlength\extrarowheight{-3pt} \begin{tabular}[r]{@{}r@{}}47.0\\ {\tiny $\pm$26.2}\end{tabular} & 
            \setlength\extrarowheight{-3pt} \begin{tabular}[r]{@{}r@{}}214.5\\ {\tiny $\pm$85.6}\end{tabular}  & 
            \setlength\extrarowheight{-3pt} \begin{tabular}[r]{@{}r@{}}42.8\\ {\tiny $\pm$33.9}\end{tabular} & 
            \setlength\extrarowheight{-3pt} \begin{tabular}[r]{@{}r@{}}128.2 \\ {\tiny $\pm$55.5}\end{tabular} & 
            \setlength\extrarowheight{-3pt} \begin{tabular}[r]{@{}r@{}}145.1\\ {\tiny $\pm$64.5}\end{tabular} & 
            
            \setlength\extrarowheight{-3pt} \begin{tabular}[r]{@{}r@{}}206.9\\ {\tiny $\pm$92.4}\end{tabular} & 
            \setlength\extrarowheight{-3pt} \begin{tabular}[r]{@{}r@{}}313.4\\ {\tiny $\pm$141.1}\end{tabular} &
            \setlength\extrarowheight{-3pt} \begin{tabular}[r]{@{}r@{}}\textbf{320.3}\\ {\tiny $\pm$159.4}\end{tabular}\\
            
 Up n Down  &
            \setlength\extrarowheight{-3pt} \begin{tabular}[r]{@{}r@{}}536.4\\{\tiny $\pm$361.5} \end{tabular}  & 
            \setlength\extrarowheight{-3pt} \begin{tabular}[r]{@{}r@{}}594.8\\ {\tiny $\pm$324.6}\end{tabular}  & 
            \setlength\extrarowheight{-3pt} \begin{tabular}[r]{@{}r@{}}823.1\\ {\tiny $\pm$320.0}\end{tabular} & 
            \setlength\extrarowheight{-3pt} \begin{tabular}[r]{@{}r@{}}348.7\\ {\tiny $\pm$227.0}\end{tabular} & 
            \setlength\extrarowheight{-3pt} \begin{tabular}[r]{@{}r@{}}641.5\\ {\tiny $\pm$428.6}\end{tabular}& 
            \setlength\extrarowheight{-3pt} \begin{tabular}[r]{@{}r@{}}383.2\\ {\tiny $\pm$242.8}\end{tabular} & 
            
            \setlength\extrarowheight{-3pt} \begin{tabular}[r]{@{}r@{}}911.5\\ {\tiny $\pm$476.7}\end{tabular} & 
            \setlength\extrarowheight{-3pt} \begin{tabular}[r]{@{}r@{}}\textbf{1072.8}\\ {\tiny $\pm$664.0}\end{tabular} &
            \setlength\extrarowheight{-3pt} \begin{tabular}[r]{@{}r@{}}990.4\\ {\tiny $\pm$707.5}\end{tabular}\\
 \bottomrule
\end{tabular}
}
\caption{
Average rewards and standard deviations (small numbers) over the last 100,000 time steps over Atari environments.
}
\label{tab:atari_result_std_summary}
\end{table*}
%%%%%%%%%%%%%%%%%%%%%%%%%%%%%%%%%%%%%%%%%%%%%%%%%%%%%%%

Interestingly, the performance of \TempoRL{} is a lot worse than the one described in the original paper~\cite{biedenkapp2021temporl}. 
It is because we use function approximation to estimate Q-values, rather than tabular Q-learning. 
Generally, uncontrolled or undesirable overestimation bias can be caused when using function approximation~\cite{moskovitz2021tactical}.
Therefore, simply selecting \rc{} with the highest value leads to a catastrophic result, especially in function approximation setting. 
Table~\ref{tab:gid_result} verifies the fact that pessimistic extension policies perform well in Lava Gridworlds.
Moreover, Table~\ref{tab:grid_tee_mu} in Appendix shows that more negative $\lambda$ achieves higher AUC scores.

\textbf{Coverage.} In Figure~\ref{fig:lava2,3_coverage}, we present coverage plots comparing \Algname{} and \TempoRL{} on two types of Lava environments. 
For \Algname{}, we have set $\lambda$ to -1.5, a value that has demonstrated robust performance across tests.
The results show that \Algname{} provides significantly better coverage over the state space.
We can induce our algorithm to repeat sub-optimal action less by using a pessimistic \repetition{} policy.
Owing to this, our agent can survive for a longer time, leading to better coverage. 

\textbf{Distribution of Extension Length.} Figure~\ref{fig:lava1_hist.pdf} depicts the \rc{} distributions of \TempoRL{} and \Algname{} on Birdge and ZigZag with logarithmically decaying $\epsilon$ exploration schedule.  
More red represents more repetitions. 
It shows that \Algname{} prefers fewer repetitions compared to \TempoRL{} when $\lambda <0$.
As previously articulated in the final paragraph in Preliminaries, observations for long extensions are seldom employed in the process of updating Q-values. Consequently, this propels our algorithm to favor fewer repetitions when $\lambda <0$.
In a pessimistic extension policy, the agent tends to refrain from repeating the chosen action many times because the value of a distant state could be much more uncertain than that of a neighbor one.

%%%%%%%%%%%%%%%%%%%%%%%%%%%%%%%%%%%%%%%%%%%%%%%%%%%%%%%%%%%%%%%%%%%%%%%%%%%%%%%%%%%%%%%%%%%

\subsection{Atari 2600: Arcade Learning Environment} \label{exp:atari}
In this section, we evaluate the performance of \Algname{} on the Atari benchmark, comparing the following six baseline algorithms: 
i) vanilla \DDQN~\cite{van2016deep}, 
ii) Fixed Repeat ($j=4$), 
iii) \TEE{}~\cite{dabney2020temporally}, 
iv) \DAR{} (Dynamic Action Repetition~\cite{lakshminarayanan2017dynamic}), 
v) \TempoRL{}~\cite{biedenkapp2021temporl}
vi) \BootstrappedDQN{} (Bootstrapped DQN)~\cite{osband2016deep}. 
The Fixed Repeat is an algorithm that naively repeats the action a fixed amount of times.

\textbf{Setup.} 
Each algorithm is trained for a total of $2.5\times10^6$ training steps, which is only 10 million frames.
All algorithms except \BootstrappedDQN{} use a linearly decaying $\epsilon$-greedy exploration schedule over the first 200,000 time-steps with a final $\epsilon$ fixed to 0.01.
We evaluated all agents every 10,000 training steps and evaluated for 3 episodes with a very small $\epsilon$ exploration rate (0.001). 
We used \textit{OpenAi Gym's} Atari environment with 4 frame-skips~\cite{bellemare2013arcade}. 
For maximal \rc{}, we set it to $10$.~\footnote{This approach aligns with the settings of \citet{biedenkapp2021temporl} to ensure a fair comparison.}

\textbf{Uncertainty-Awareness.} 
Figure~\ref{fig:atari_best4_learning_curve} depicts learning curves for \Algname{} and other baseline algorithms (see Figure~\ref{fig:atari_learning_curve} for full version).
And Table~\ref{tab:atari_result_std_summary} summarizes the results of the games in terms of average rewards over the last 100,000 time steps (refer Table~\ref{tab:atari_result_std} for other environments).
Overall, \Algname{} achieves higher final rewards than other agents.
These results demonstrate that if $\lambda$ is properly tuned to the environment, our method shows significantly improved performance than existing action repetition methods (\DAR{}, \TEE{} and \TempoRL{}) as well as a deep exploration algorithm (\BootstrappedDQN{}).
On top of that, we found that the Fixed Repeat algorithm fails at learning in most games.
Hence, it is crucial to learn a \repetition{} policy for higher performance.

\textbf{Effect of $n$-step Learning.} 
We investigate the effect of $n$-step \Q{}-learning for the action-value function $\behaviorQ{}$, and our empirical results suggest that it helps in most games.
Table~\ref{tab:atari_result_std_summary} shows that applying $n$-step learning improves performance remarkably.
We present the extended version of the ablation study in Figure~\ref{fig:ablation_nstep}, which shows a $30.2\%$ improvement (from 1.39 to 1.81) after $n$-step Learning has been applied.
The result empirically supports our previous argument that off-policy correction is not necessary for our action-repeating options framework.

\textbf{Adaptive Uncertainty Parameter $\lambda$.}
As illustrated in Figure~\ref{fig:atari_best4_learning_curve} and Table~\ref{tab:atari_result_std_summary}, the learning speed of \Algname{} with adaptively chosen $\lambda$ is somewhat slower compared to the standard \Algname{}. This slight decrease in speed primarily stems from the need for additional samples to optimize $\lambda$.
Nevertheless, even with this adjustment, \Algname{} with an adaptive $\lambda$ continues to outperform other baseline methods by a considerable margin. 
These results are particularly encouraging as they obviate the need to predefine the value of $\lambda$, thereby reducing the burden of hyperparameter tuning. 
This aspect of our approach further underscores its practicality and effectiveness in complex learning scenarios.

%%%%%%%%%%%%%%%%%%%%%%%%%%%%%%%%%%%%%%%%%%%%%%%%%%%%%%%%%%%%%%%%%%%%%%%%%%%%%%%%%%%%%%%%%%%
\subsection{Control Problem: Pendulum-v0} \label{exp:adversary}
In this section, we show that \Algname{} maintains its robustness to continuous control problems where there is a significant chance that repeated actions will surpass the balancing point. 
Consequently, selecting the appropriate \rc{} becomes even more crucial.
We choose to evaluate on OpenAI gyms~\citep{brockman2016openai} Pendulum-v0.
Since the action space is continuous, we use \texttt{DDPG}~\citep{lillicrap2015continuous} as our action policy $\actionpolicy$, thus label it as \texttt{UTE-DDPG}, and apply the adaptive uncertainty parameter technique.
The baseline agents are \texttt{DDPG}~\citep{lillicrap2015continuous}, \texttt{FiGAR}~\citep{sharma2017learning}, and \texttt{t-DDPG} (\TempoRL{}-\texttt{DDPG})~\citep{biedenkapp2021temporl}.

\textbf{Setup.} 
We trained all agents for a total of $3 \times 10^4$ training steps with evaluations conducted every $250$ steps. 
For the initial $10^3$ steps, a uniform random policy was applied to accumulate initial experiences. 

\textbf{Robustness to Continuous Control Environment.} 
In Table~\ref{tab:pendulum}, \texttt{UTE-DDPG} (with adaptively chosen $\lambda$) demonstrates superior performance, achieving either the top or second-best performance among the benchmarks. 
This suggests that our algorithm is robust to continuous control environments and consistently outperforms other established action-repeating algorithms, such as \texttt{FiGAR} and \texttt{t-DDPG}. 
This advantage can be attributed to our \textit{uncertainty-aware} extension policy that prudently repeats actions. 
Additionally, \Algname{} exhibits smaller standard deviations than all other baselines, except when the maximal \rc{} is large (i.e., $J=8$), indicating enhanced learning stability.

\begin{table}[ht]
\centering 
\begin{tabular}{c  | r  rrr} 
\toprule
Max $J$ & \texttt{DDPG} & \texttt{FiGAR}   & \texttt{t-DDPG} & \texttt{UTE-DDPG} \\
\midrule
2                         & \multirow{7}{*}{\setlength\extrarowheight{-3pt} \begin{tabular}[r]{@{}r@{}}\textbf{-156.9}\\{\tiny $\pm$23.2} \end{tabular}}  
                          & \begin{tabular}[r]{@{}r@{}}-172.7\\{\tiny $\pm$48.6} \end{tabular} 
                          & \begin{tabular}[r]{@{}r@{}}-163.2\\{\tiny $\pm$28.6} \end{tabular} 
                          & \begin{tabular}[r]{@{}r@{}}\textbf{-152.6}\\{\tiny $\pm$17.2} \end{tabular}   
                          \\
4                         &                          
                          & \begin{tabular}[r]{@{}r@{}}-352.8 \\{\tiny $\pm$181.5} \end{tabular} 
                          & \begin{tabular}[r]{@{}r@{}}-160.1 \\{\tiny $\pm$50.7} \end{tabular}
                          & \begin{tabular}[r]{@{}r@{}}\textbf{-147.4}\\{\tiny $\pm$17.1} \end{tabular}              
                          \\
6                         &                          
                          & \begin{tabular}[r]{@{}r@{}}-831.2 \\{\tiny $\pm$427.0} \end{tabular} 
                          & \begin{tabular}[r]{@{}r@{}}-163.5\\{\tiny $\pm$29.0} \end{tabular} 
                          & \begin{tabular}[r]{@{}r@{}}\textbf{-159.2}\\{\tiny $\pm$20.8} \end{tabular}              
                          \\
8                         &                          
                          & \begin{tabular}[r]{@{}r@{}}-1295.0\\{\tiny $\pm$274.5} \end{tabular} 
                          & \begin{tabular}[r]{@{}r@{}}-175.3\\{\tiny $\pm$60.3} \end{tabular} 
                          & \begin{tabular}[r]{@{}r@{}}\textbf{-165.0}\\{\tiny $\pm$26.4} \end{tabular}        
                          \\
\bottomrule
\end{tabular}%
\caption{
Average rewards and standard deviations (small numbers) over the last 10,000 time steps in Pendulum-v0 over various maximal \rc{}s ($J$).
}
\label{tab:pendulum}
\end{table}
% 

%%%%%%%%%%%%%%%%%%%%%%%%%%%%%%%%%%%%%%%%%%%%%%%%%%%%%%%%%%%%%%%%%%%%%%%%%%%%%%%%%%%%%%%%%%%
\section{Conclusion}
\label{sec6:conclusion}
We propose a novel method that learns to repeat actions while explicitly considering the uncertainty over the Q-value estimates of the states reached under the repeated-action option.
By calibrating the level of uncertainty considered (denoted by $\lambda$), \Algname{} consistently and significantly outperforms other algorithms, especially those focusing on action repetition, across various environments such as Chain MDP, Gridworlds, Atari 2600, and even in control problems.
To our best knowledge, this is the first deep RL algorithm considering uncertainty in the future when instantiating temporally extended actions.

%%%%%%%%%%%%%%%%%%%%%%%%%%%%%%%%%%%%%%%%%%%%%%%%%%%%%%%%%%%%%%%%%%%%%%%%%%%%%%%%%%%%%%%%%%%

\section*{Acknowledgments}

This work was supported 
by Creative-Pioneering Researchers Program through Seoul National University, and by
the National Research Foundation of Korea (NRF) grant funded by the Korea government (MSIT) (No.
2022R1C1C100685912, 
2022R1A4A103057912, 
and RS-2023-00222663).

\bibliography{aaai24}

% APPENDIX
%%%%%%%%%%%%%%%%%%%%%%%%%%%%%%%%%%%%%%%%%%%%%%%%%%%%%%%%%%%%%%%%%%%%%%%%%%%%%%%
%%%%%%%%%%%%%%%%%%%%%%%%%%%%%%%%%%%%%%%%%%%%%%%%%%%%%%%%%%%%%%%%%%%%%%%%%%%%%%%
\clearpage

\appendix
\onecolumn
% In the unusual situation where you want a paper to appear in the
% references without citing it in the main text, use \nocite
% \nocite{langley00}
% \counterwithin{figure}{section}
% \counterwithin{table}{section}
% \counterwithin{equation}{section}
%%%%%%%%%%%%%%%%%%%%%%%%%%%%%%%%%%%%%%%%%%%%%%%%%%%%%%%%%%%%%%%%%%%%%%%%%%%%%%%%%%%%%%%%%%%

\section{Details of Baselines}\label{app:baselines} 

\textbf{Fixed Repeat.}
Fixed Repeat in Atari experiment corresponds to a \DDQN{} agent that always repeats the action for a fixed amount of times. 
In other words, the extension policy returns the same $j$ at every decision time.
In our settings, $j$ is set to 4 (see Figure~\ref{fig:appendix_fixed_j} for performance of other $j$s).
The reason for evaluating this naive method is to confirm that this approach fails, highlighting the importance of the \rc{}.

\textbf{Temporally-Extended $\epsilon$-Greedy.}
\TEE{}~\cite{dabney2020temporally} is a simple add-on to the $\epsilon$-greedy policy. 
The agent follows the current policy for one step with probability $1-\epsilon$, or with probability $\epsilon$ samples an action $a$ from a uniform random distribution and repeats it for $j$ times, which is drawn from a pre-defined duration distribution.
We used the heavy-tailed zeta distribution, with $\mu=1.25$ as the duration distribution in the Chain MDP and the Gridworlds environment. This was done by conducting a hyperparameter search on $\mu$ for the set \{1.25, 1.5, 2.0, 2.5, 3.0\}.
In Atari games, we choose the best per-game $\mu$ among the set \{1.5, 1.75, 2.0, 2.25, 2.5\} for a fair comparison to our \Algname{}.
A combination of $\epsilon$ chance to explore and zeta-distributed duration is called $\epsilon z$-greedy exploration.

The experimental results from ~\cite{dabney2020temporally} show that \TEE{} incorporated in existing R2D2 and Rainbow agents result higher median human-normalized score over the 57 Atari games.
However, this algorithm is highly dependent on the exploration rate $\epsilon$, which can cause difficulties in online learning.

\textbf{Dynamic Action Repetition.} 
\DAR{}~\cite{lakshminarayanan2017dynamic} is a framework for discrete-action space deep RL algorithms. 
\DAR{} duplicates the output heads twice such that an agent can choose from $2\times |A|$ actions.
And each output heads corresponds to pre-defined repetition values, $r_1, r_2$, where $r_1$ and $r_2$ are fixed hyper parameters.
Hence, action $a_{k}$ is repeated $r_1$ number of times if $k < |\mathcal{A}|$ and $r_2$ number of times if $k \ge | \mathcal{A}|$. 
In our experiments, $r_1$ is fixed to maximum \rc{} $J$ and $r_2$ to 1 to allow for actions at every time step.

There are some drawbacks to this approach.
First, $r_1$ and $r_2$ have to be predefined, which means we need prior knowledge of the environments. 
And also, the learning process becomes a lot more difficult because the action space doubled.

\textbf{TempoRL.}
\TempoRL{}~\cite{biedenkapp2021temporl} proposes a ``flat'' hierarchical structure in which \textit{behavior} policy ($\pi_a$) determines the action $a$ to be played given the current state $s$, and a \textit{skip} policy ($\pi_j$) determines how long to repeat this action.
The flat hierarchical structure refers to \textit{behavior} policy and \textit{skip} policy having to make decisions at the same time-step. The action policy has to be always queried before the skip policy.
When an agent plays a chosen action for \rc{} $j$, total of $\frac{j\dot(j+1)}{2}$ skip-transitions are observed and stored in the replay buffer. The \textit{behavior} and the \textit{skip} Q-functions can be updated using one-step observations and the overarching skip-observation. 
Using the samples collected, the \textit{behavior} policy can be learned by a classical one step \Q{}-learning.
The n-step Q-learning is used to learn the \textit{skip} value with the condition that, at each step in the $j$ steps, the action stays the same.

\textbf{Bootstrapped DQN.}
\BootstrappedDQN{}~\cite{osband2016deep} is an algorithm for temporally-extended (or deep) exploration. Inspired by Thompson sampling, it selects an action without the need for an intractable exact posterior update. \citet{osband2016deep} suggest bootstrapped neural nets can produce reasonable posterior estimates. The network of bootstrapped DQN consists of a shared architecture with $K$ bootstrapped ``heads'' stretching off independently. Each head is initialized randomly and trained only on its bootstrapped sub-sample of the data. The shared network learns a joint feature representation across all the data. For evaluation, an ensemble voting policy is used to decide action.

\textbf{FiGAR.}
\texttt{FiGAR}~\citep{sharma2017learning} is a framework tailored for both discrete and continuous action spaces. 
Unlike the \DAR{} method where a single policy learns both the action selection and its duration, \texttt{FiGAR} separates these tasks using two distinct policies: $\actionpolicy: \mathcal{S} \rightarrow \mathcal{A}$ for action selection and $\pi_r: \mathcal{S} \rightarrow  \{1, 2, ..., \text{max repetition}\}$ for determining repetition duration. 
During the training process, given a state $s$, $\actionpolicy$ chooses the action while $\pi_r$ simultaneously determines how long that action should be repeated starting from $s$. 
Crucially, when making their selections, neither $\actionpolicy$ nor $\pi_r$ has knowledge of the other's decision. 
This ensures that the action and its repetition duration are chosen independently.

\section{Implementation Details: \Algname} \label{app:Implementation_details}
\subsection{Bootstrap with random initialization for option-value functions} Formally, we consider an ensemble of $B$ option-value functions, $\{\Tilde{Q}^{\policyoveroption{}}_{(b)}\}_{b=1}^B$, where $\policyoveroption{}$ denotes the 
policy over option. 
To train the ensemble of option-value functions $\Tilde{Q}^{\policyoveroption{}}_{(b)}$, we use two mechanisms to enforce diversity between these \Q{}-functions~\cite{efron1982jackknife, osband2016deep}: The first mechanism is random-initialization of model parameters for each option-value functions to induce initial diversity in the models. 
The second mechanism is to train each \Q{}-function with different samples.
Specifically, in each timestep $t$, each $b^{th}$ \Q{}-function  is trained by multiplying binary mask $m_{t,b}$ to each objective function, where the binary mask $m_{t,b}$ is sampled from the Bernoulli distribution ($p$) with parameter $\beta\in(0,1]$.
In our experiments, we use $p=0.5$ for the parameter of the Bernoulli distribution.

\subsection{Multi-step target for both action- and option-value functions} 
We learn parameterized estimates of Q-value functions, an action-value function $\behaviorQparam{} \approx \optimalbehaviorQ{}(s,a)$ and an option-value function $\skipQparam{} \approx \optimalskipQ{}(s,\omega_{aj})$, using neural networks.
We use two different neural network function approximators parameterized by $\theta$ and $\phi$ respectively.
For stability, we integrate double \Q-learning \cite{van2016deep} technique.
We use multi-step \Q-learning to update both action-value function $\behaviorQ{}$ and option-value function $\skipQ{}$.
The following equations represent Bellman residual errors of action- and option-value functions respectively:
\begin{align}
 \mathcal{L}_{\behaviorQ{}}(\theta) &= \mathbb{E}_{\tau_{t} \sim \mathcal{R}} \bigg[(\Q^{\pi_{\omega}}(s_t, a_t ; \theta) - \sum_{k=0}^{n-1}\gamma^k r_{t} - \gamma^n \max_{a'}\,\optimalbehaviorQ{}(s_{t+n},a'; \Bar{\theta})  )^2 \,\biggl\vert\, n \leq j \sim \extensionpolicy{} \bigg] \label{eq:action_value_loss} \\
  \mathcal{L}_{\skipQ{}}(\phi) &= \mathbb{E}_{\tau_{t} \sim \mathcal{R}} \bigg[  \sum_{b=1}^{B}\, \Big[ m_{t,b} (\skipQb{}(s_t, \omega_{aj} ; \phi) - \sum_{k=0}^{j-1}\gamma^k r_{t} - \gamma^j \max_{a'}\,\optimalbehaviorQ{}(s_{t+j},a'; \Bar{\theta})  )^2 \Big] \bigg] \label{eq:option_value_loss}
\end{align}
where $\tau_{t}$ is a multi-step transition trajectory sampled from a replay buffer $\mathcal{R}$, $m_{t,b}$ is a binary bootstrap mask, and $\bar{\theta}$ are the delayed parameters of action-value function $\behaviorQ{}$. 
The delayed parameters are the parameters of the target network for action-value function $\behaviorQ{}$.
The target network with the delayed parameters is the same as the online network except that its parameters are copied every $\tau$ step from the online network, and kept fixed on all other steps~\cite{mnih2015human}.
Note that $n$ for $n$-step learning in Eq.\eqref{eq:action_value_loss} has to follow the current extension policy $\extensionpolicy{}$. 
Therefore, in order to update action-value function $\behaviorQ{}$ in Eq.\eqref{eq:action_value_loss}, we only use trajectory samples in which the \rc{} is smaller than or equal to the output of current extension policy, i.e. $n \leq j \sim \extensionpolicy{}(j_t\mid s_t, a_t)$.

One interesting point of the above equations is that we use the same target value for both \Q{}-functions: $\behaviorQ{}$ in Eq.\eqref{eq:action_value_loss} and $\skipQ{}$ Eq.\eqref{eq:option_value_loss}.
Trivially, we can demonstrate that the target value for the option selection of repeated actions is the same as one for single-step action selection within the option, i.e. $\underset{a'}{\max}\,\optimalbehaviorQ{}(s_{t+j},a') = \underset{\omega'}{\max}\,\optimalbehaviorQ{}(s_{t+j},\omega'_{aj})$.
By using the same target value, we can stabilize the learning process.

\subsection{The Same Target for both action- and option- value functions} \label{proof_of_proposition1}
On the fourth page of the main paper, we argue that we can use the same target for both types of $\Q{}$-values.
The following proposition formalizes the statement.
Though it is a trivial result, we simply present the proof of it for better comprehension.
\begin{proof}[Proof of Propositon~\ref{prop:optimal_value}]
    For any $s \in \mathcal{S}$, define the value of executing an action in the context of a state-option pair as $\Q_{U} : \mathcal{S} \times \Omega \times \mathcal{A} \rightarrow \mathbb{R}$.
    Let $\pi_U(a \mid s, \omega_{aj}) : \mathcal{S} \times \Omega \rightarrow \mathcal{P}(\mathcal{A}) $ be an \textit{intra-option} policy~\cite{sutton1999between}, which returns an action $a$ when executing an option $\omega_{aj}$ at state $s$.
    Then, option-value functions can be written as:
    \begin{align}
        \skipQ{}(s,\omega_{aj}) &= \underset{a'}{\sum}\, \pi_U(a'|s,\omega_{aj}) \Q{}_{U}(s, \omega_{aj}, a')\nonumber 
        = \underset{a'}{\sum}\, \bm{1}_{a} \Q{}_{U}(s, \omega_{aj}, a')\nonumber\\
                                &= \Q{}_{U}(s, \omega_{aj}, a),  \label{eq:q_equal}
    \end{align}
    where the second equality holds since $\pi_U$ deterministically returns action $a$.
    Therefore we have, 
    \begin{align*}
        V^{\pi_{\omega}^*}(s_0) &= \max_{\omega_{aj}}  \Tilde{Q}^{\pi_{\omega}^*} (s_0,\omega_{aj}) = \max_{a,j}\Tilde{Q}^{\pi_{\omega}^*} (s_0,\omega_{aj})    
        = \max_{a,j} \Q^*_{U}(s, \omega_{aj}, a)   
        \\&= \max_{a}\left\{ \max_{j} \Q^*_{U}(s, \omega_{aj}, a)  \right\} 
        = \max_{a} \Q^{\pi_{\omega}^*}(s,a),
    \end{align*}
    where the second equality holds since $\omega_{aj}$ is determined by an action $a$ and \rc{} $j$, the third equality is by Eq.\eqref{eq:q_equal}, and the last equality holds since $\pi_{\omega}^*$ is the optimal policy. 
    This concludes the proof.
\end{proof}

\section{Experiments Details}\label{app:experiments_details}
All experiments were run on an internal cluster containing GeForce RTX 3090 GPUs. Atari experiments took 13 hours to train for 10 million frames on GPU.
Our Chain MDP environment and \BootstrappedDQN{} baseline implementation is based on code from~\citet{touati2020randomized}. The license for this asset is Attribution-NonCommercial 4.0 International.
Gridworlds and Atari environment settings along with \TempoRL{} baseline implementation is from~\citet{biedenkapp2021temporl}. This asset is licensed under Apache License 2.0. The Arcade Learning Environment (ALE)~\cite{bellemare2013arcade} for Atari games is licensed under the GNU General Public License Version 2. 

\subsection{Chain MDP experiment}
\textbf{Network Architecture.} In the Chain MDP experiment, all agents, i.e. \TEE{}, \TempoRL{}, and \Algname{}, use simple DQN architecture~\cite{mnih2015human} for their action policy. 
The network consists of 3 dense layers with a ReLU activation function.
The number of hidden nodes is set to 16 for all dense layers.\\
\TempoRL{} has another output stream that combines a hidden layer with 10 units together with the output of the second fully connected layer. 
It is followed by a fully connected layer that outputs predicted Q-values, \rc{}s. \\
In order to implement \repetition{} policy $\extensionpolicy{}$ of \Algname{}, the agent has another ensemble network of 10 identical neural networks.
Each of these 10 ensemble networks is a 3-layer neural network with fully connected layers with 26 hidden units, where the input is a concatenation of the state and the chosen action.

\begin{table*}[ht]
\centering
    \begin{tabular}{l|c}
    \toprule
        Hyper-parameter & Value  \\
    \midrule
    Discount rate  & 0.999  \\
    Target update frequency & 500 \\
    Initial $\epsilon$ & 1.0 \\
    Final $\epsilon$ & 0.001 \\
    $\epsilon$ time-steps & N $\times$ 100 \\
    Loss Function & Huber Loss \\
    Optimizer & Adam \\
    Learning rate & 0.0005 \\
    Batch Size & 64 \\
    Replay buffer size    & $5 \times 10^4$  \\
    Extension replay buffer size & $5 \times 10^4$ \\
    Number of ensemble heads & 10 \\
    Max \rc{} ($J$) & 10, 15, 20 \\
    Uncertainty parameter ($\lambda$) & -2, -1, 0, 1, 2 \\
    \bottomrule
    \end{tabular}%
\label{tab:nchain_hyperparams}
\caption{Hyper-parameters used for the Chain MDP experiments}
\end{table*}

\subsection{Gridworlds experiment}
\textbf{Network Architecture.} In the Gridworlds experiment, all agents, i.e. \DDQN{}, \TempoRL{} and \Algname{}, were trained using deep Q-network \cite{mnih2015human} with 3 dense layers.
The number of each hidden node is 50 and ReLU activation function was used for non-linearity. 
All of the agents were implemented using double DQN \cite{van2016deep}.
Both \TempoRL{} and \Algname{} agents have separate network for \repetition{} policy.

For \repetition{} policy, \TempoRL{} uses a single 3-layer neural network with fully connected layers of 50, 50, and 50 units, whereas our \Algname{} uses 10 duplicated networks of a 3-layer neural network with fully connected layers of 50, 50, and 50 units.
The input of \repetition{} policy is a concatenation of the state and the chosen action, which is the same as the Chain MDP experiment.

\begin{table}[ht]
\centering
    \begin{tabular}{l|c}
    \toprule
        Hyper-parameter & Value  \\
    \midrule
    Discount rate  & 0.99  \\
    Initial $\epsilon$ & 1.0 \\
    Final $\epsilon$ & 0.0 \\
    $\epsilon$ time-steps & 50 \\
    Loss Function & MSE Loss \\
    Optimizer & Adam \\
    Learning rate & 0.001 \\
    Batch Size & 64 \\
    Replay buffer size    & $10^6$  \\
    Extension replay buffer size & $10^6$ \\
    Number of ensemble heads & 10 \\
    Max \rc{} ($J$) & 7 \\
    Uncertainty parameter ($\lambda$) & -1.5, -1.0, -0.5 \\
    \bottomrule
    \end{tabular}%
\label{tab:gird_hyperparams}
\caption{Hyper-parameters used for the Gridworlds experiments}
\end{table}

\subsection{Atari experiment} \label{app:atari_details}
\textbf{Network Architecture.} The input size of images is $84 \times 84$, and the last 4 frames of this image are stacked together.
This will be our input throughout the experiment.

\begin{table}[ht]
\centering
    \begin{tabular}{l|c}
    % \vspace{+10pt}
    \toprule
        Hyper-parameter & Value  \\
    \midrule
    Discount rate  & 0.99  \\
    Gradient Clip & 40.0 \\
    Target update frequency & 500 \\
    Learning starts & 10 000 \\
    Initial $\epsilon$ & 1.0 \\
    Final $\epsilon$ & 0.01 \\
    Evaluation $\epsilon$ & 0.001 \\
    $\epsilon$ time-steps & 200 000 \\
    Train frequency & 4 \\
    Loss Function & Huber Loss \\
    Optimizer & Adam \\
    Learning rate & 0.0001 \\
    Batch Size & 32 \\
    Extension Batch Size & 32 \\
    Replay buffer size    & $5 \times 10^4$  \\
    Extension replay buffer size & $5 \times 10^4$ \\
    Number of ensemble heads & 10 \\
    Max \rc{} ($J$) & 10 \\
    Uncertainty parameter ($\lambda$) & -1.5, -1.0, -0.5, -0.2, \\
                               &  0.0, 0.2, 0.5, 1.0 \\
    \bottomrule
    \end{tabular}%
    \label{tab:atari_hyperparams}
\caption{Hyper-parameters used for the Atari experiments}
\end{table}

\textbf{\DDQN{}} agent uses the same architecture for DQN of~\cite{mnih2015human} with the target network~\cite{van2016deep}.
This architecture has 3 convolutional layers of 32, 64 and 64 feature planes with kernel sizes of 8,4 and 3, and strides of 4,2, and 1, respectively.
These are followed by a fully connected network with 512 hidden units followed by another fully connected layer to the \Q-Values for each action.\\
\textbf{\TEE{}} agent uses the exact same architecture as~\citet{mnih2015human}. 
The only difference with \DDQN{} is that \TEE{} repeats an exploratory action, which is sampled from uniform random distribution.
And the \rc{} $j$ is sampled from zeta distribution.
The hyper-parameter $\mu$ for zeta distribution is set depending on the experiments: 1.25 for Chain-MDP and Gridwolrds, and the best one for each game in Atari experiment.  \\
\textbf{\DAR{}} agent selects action and \rc{} based on $2 \times |\mathcal{A}|$ Q-values. Therefore, the output of the last layer is duplicated and the duplicate outputs corresponding to a different \rc{}, $r_1$ and $r_2$.
The hyper-parameters, $r_1$ and $r_2$, are set to 1 and 10 respectively. \\
\textbf{\TempoRL{}} agent uses the shared architecture, the structure of which is the same as one described in~\citet{biedenkapp2021temporl}.
On top of DQN architecture~\cite{mnih2015human}, an additional output stream for the \rc{} is incorporated. 
The \rc{} is embedded into a 10-dimensional vector and then concatenated with the output of the last convolutional layer of the network.
The features then pass through two fully connected hidden layers, each with 512 units. \\
\textbf{\BootstrappedDQN{}} has one torso network of 3 convolutional layers, which is the same as that of DQN~\cite{mnih2015human}. 
However, it has 10 heads branching off independently~\cite{osband2016deep}. 
Each head consists of two fully connected hidden layers, each with 512 units.
Therefore the agent returns 10 \Q-values from each head.\\
\textbf{\Algname{}} uses the similar architecture as that of \TempoRL{}~\cite{biedenkapp2021temporl}.
The main difference compared to \TempoRL{} is that \Algname{} uses an ensemble method for the output stream of \rc{}.
After concatenating a 10-dimensional \rc{} vector and the output of the last convolutional layer, the concatenated vector pass through 10 heads branching off independently.
Each 10 head consists of fully connected layer with 512 hidden units followed by a fully connected layer to the \Q-Values for each \rc{}.

%%%%%%%%%%%%%%%%%%%%%%%%% Nchain %%%%%%%%%%%%%%%%%%%%%%%%%%%%%

\begin{figure}[ht]  %[htp!]
    \includegraphics[width=\textwidth]{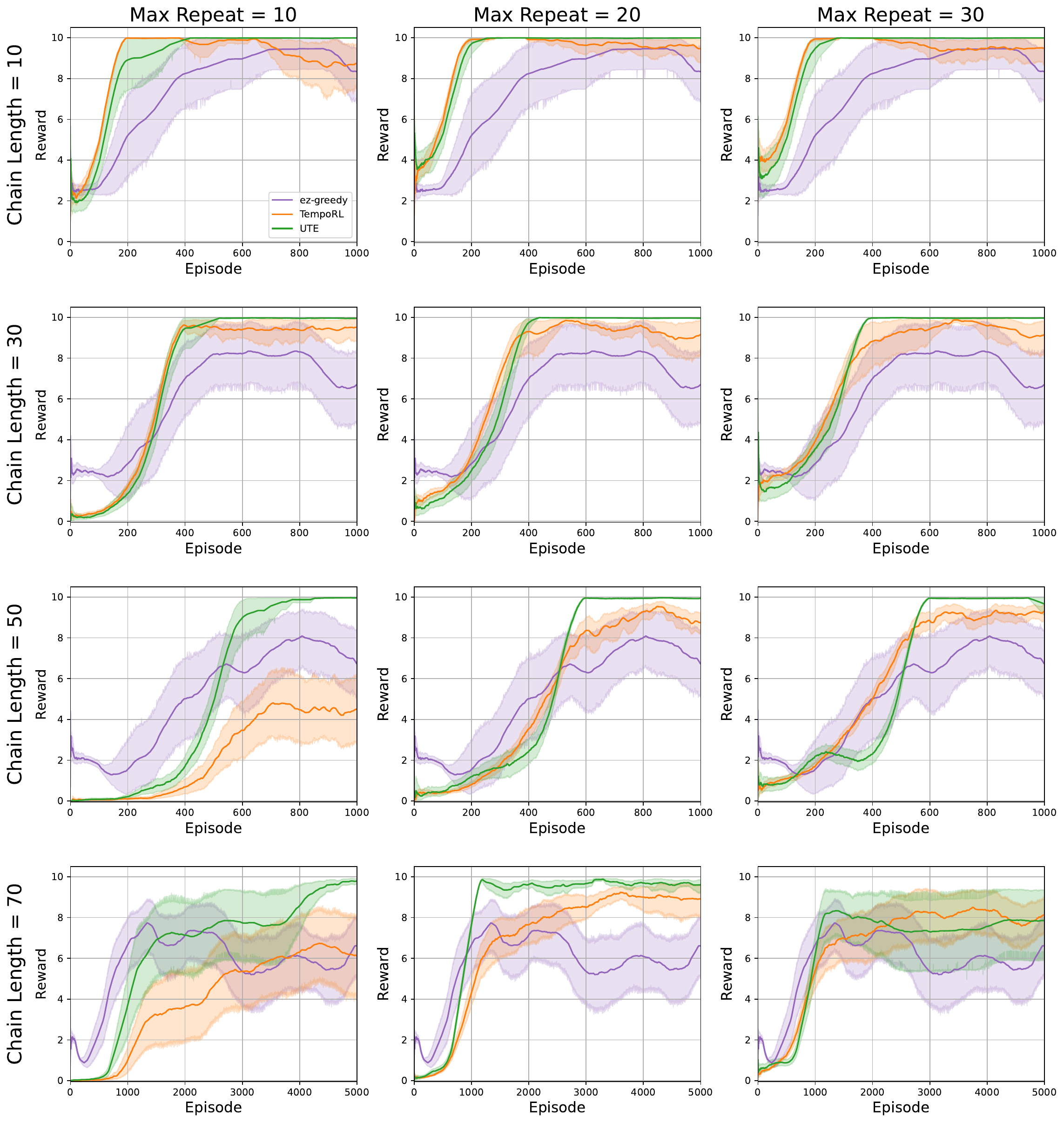}
    \centering
    
    \caption{Training reward for \TEE{}, \TempoRL{}, and \Algname{} in the Chain MDP environment. 
    The horizontal axis represents the maximum \rc{} of 10, 20, and 30 from left to right, respectively. The vertical axis represents the chain length of 10, 30, 50, and 70 from top to bottom. Learning curve for $N = 70$ is presented with 5,000 training episodes. (20 random seeds)}
    \label{fig:ncahin_performance_all}
\end{figure}

\section{Further Experimental Results}\label{app:further_results}
\subsection{Chain MDP}

\textbf{Uncertainty Parameter.} Table~\ref{tab:nchain_auc_all} shows the effect of the \riskparameter{} on normalized AUC score for 1,000 training episodes. 
We can see that an exploration-favoring high \riskparameter{} is beneficial in the Chain MDP environment.
The longer the chain length, the more sensitive it becomes sensitive to the uncertainty parameter.
In the chain length of 70, \Algname{} with uncertainty parameter +2 is a lot better than the one with uncertainty parameter -2.
This result indicates that optimistically repeating the chosen action could lead to good performance if there is no risky area in the environment.
The Table also shows that for $\TEE{}$ with $\mu=1.25$ performed the best, and we used the value for the experiments.

\textbf{Maximum Repeat $J$.} We can also predispose the agent to repeat actions in larger numbers by increasing another parameter, the maximum \rc{} $J$.  
However, increasing the maximum \rc{} is not always a good solution as it increases the size of the set of \rc{}s, $|\mathcal{J}|$, slowing down the learning process. 
As exhibited in Figure~\ref{fig:ncahin_performance_all}, the small maximum \rc{} is detrimental to the agent's performance.
Note that the degree of exploration of \TEE{} is affected by the hyperparameter $\mu$ for zeta distribution, and the value is fixed to 1.25 throughout the chain MDP experiments.
While \TempoRL{} is sensitive to \rc{} especially when the chain length is long, our \Algname{} is quite robust to changes in \rc{}.
We can see that \Algname{} agent reaches the highest final performance compared to other agents. 
\begin{table*}[ht]
\centering
    \begin{tabular}{c|ccccccccccc}
    \toprule
                 & \multicolumn{1}{c}{\multirow{2}{*}{\TempoRL}} & \multicolumn{5}{c}{\TEE{} ($\mu$)}                        & \multicolumn{5}{c}{\Algname{} ($\lambda$)} \\
    Chain Length & \multicolumn{1}{c|}{}                         & 1.25  & 1.5 & 2.0 & 2.5 & \multicolumn{1}{l|}{3.0} & -2.0  & -1.0  & 0.0  & 1.0  & 2.0 \\ 
    \midrule
    10           & \multicolumn{1}{l|}{0.90}                   & \textbf{0.65} & 0.61 & 0.48  &  0.46   & \multicolumn{1}{l|}{0.46}  & 0.88    & 0.91    & 0.90   & 0.92   & \textbf{0.92}  \\
    30           & \multicolumn{1}{l|}{0.74}                    & \textbf{0.43} &0.42 & 0.42  &  0.38   & \multicolumn{1}{l|}{0.27}  &  0.45   & 0.55    & 0.62   & 0.73   & \textbf{0.76}  \\
    50           & \multicolumn{1}{l|}{0.25}                   & \textbf{0.43} & 0.40    & 0.23  & 0.17    & \multicolumn{1}{l|}{0.05}  & 0.07     & 0.05    & 0.37    & 0.47   &  \textbf{0.67} \\
    70           & \multicolumn{1}{l|}{0.05}                   & \textbf{0.13} & 0.12    & 0.03  & 0.04    & \multicolumn{1}{l|}{0.1}  & 0.01    &  0.01   & 0.01   &    0.07& \textbf{0.19}  \\
    \bottomrule
    \end{tabular}
\caption{
Normalized AUC for rewar across different agents over 20 random seeds. Maximum \rc{} for \TempoRL{} and \Algname{} is set to 10. The numbers in the brackets of \Algname{} represent the \riskparameter{}, $\lambda$. We can see that greater \riskparameter{}s show better performance, especially in the difficult settings where the chain length is long.
}
\label{tab:nchain_auc_all}
\end{table*}

%%%%%%%%%%%%%%%%%%%%%%%%% Grid %%%%%%%%%%%%%%%%%%%%%%%%%%%%%
\subsection{Gridworlds}
In the following additional results, we supplement one more environment called Cliff in addition to Bridge and Zigzag.
Same as others, the Cliff is discrete, deterministic, and has a $6 \times 10$ size with sparse rewards.
Note that all agents are implemented by function approximation, not tabular setting.

\begin{figure}[h]
    \centering
        \includegraphics[clip, trim=3cm 0cm 3cm 0.0cm, width=0.85\textwidth]{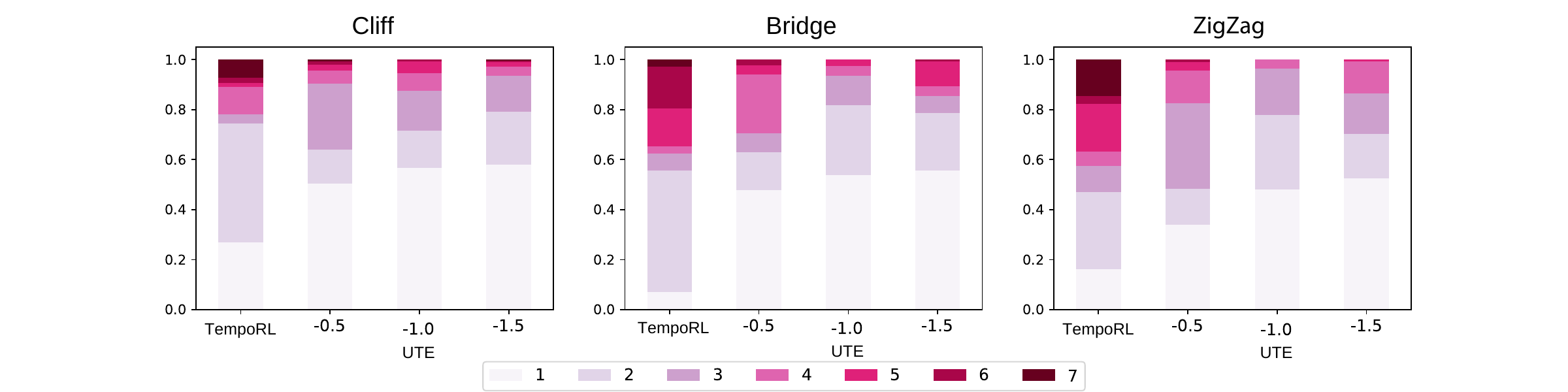}  %trim = left, botm, right, top
    \caption{Distribution of \rc{} for three Lava environments with logarithmically decaying $\epsilon$ exploration schedule. More red represents more repetitions.}
    \label{fig:grid_lava_total_log_skip_hist}
\end{figure}

\textbf{Uncertainty Parameter}
Table~\ref{tab:grid_tee_mu} shows an uncertainty-averse (negative \riskparameter{}) is beneficial in the Gridworlds environment where there are risky area (Lava).
Combined with the result of Figure~\ref{fig:grid_lava_total_log_skip_hist}, we empirically verified that more negative $\lambda$ induce more pessimistic behavior, which can lead to better performance.
The Table also shows that for $\TEE{}$ with $\mu=1.25$ performed good in overall, and we used the value for the experiments.

\begin{figure}[ht]
    \centering
        \includegraphics[clip, trim=2.7cm 2.0cm 2.7cm 2.0cm, width=0.85\textwidth]{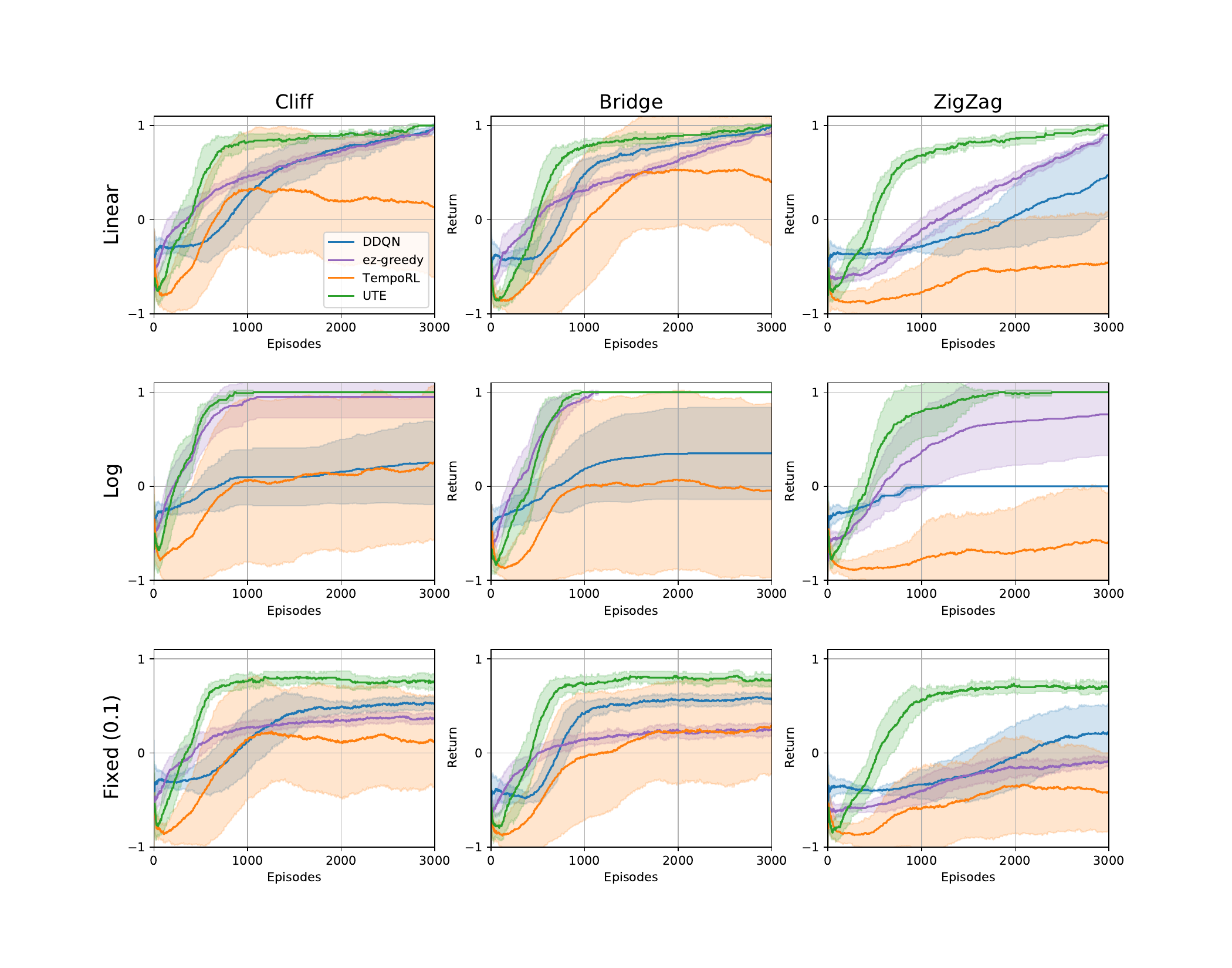}  %trim = left, botm, right, top
    \caption{Learning curves across three Lava environments and three different $\epsilon$-decaying exploration strategies, comparing \Algname{} with  \DDQN{}, \TEE{}  and \TempoRL{}. 
    Shaded areas represent the standard deviations over 20 random seeds.}
    \label{fig:grid_learning_curve}
\end{figure}

\textbf{Learning curve.} In Figure~\ref{fig:grid_learning_curve}, we plot all the learning curves of three agents across three different Lava environments and three different exploration schedules. 
In this result, we observe that \Algname{} converges to the optimal solution faster than other baselines, showing low standard deviations.
Also, our \Algname{} is robust to varying exploration strategies even using sub-optimal ones such as Fixed~$\epsilon$.
One more interesting point is that \TempoRL{} performs worse than vanilla \DDQN{}.
These results contradict the ones of~\citet{biedenkapp2021temporl}, which experimented in a tabular setting instead of function approximation.
Generally, when using function approximation to estimate \Q{}-values, it is more likely to choose sub-optimal action $a$.
Therefore, in this case, it is necessary to consider uncertainty in estimated values for safely repeating the chosen action.
By inducing pessimism ($\lambda < 0$) to the \repetition{} policy $\extensionpolicy{}$, the agent can repeat the chosen sub-optimal action $a$ less, which leads to a safer learning.

\begin{figure}[ht] %[htp!]
    \includegraphics[clip, trim=0.0cm 7.2cm 11.0cm 2.2cm, width=0.85\textwidth]{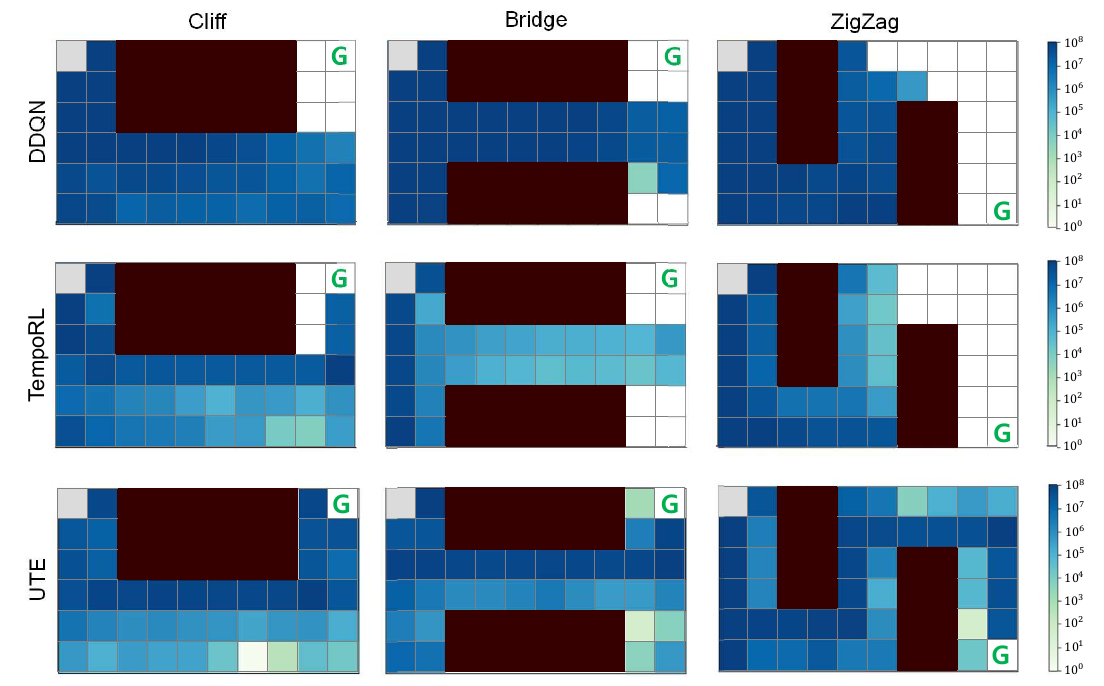}
    \centering
    \caption{Coverage plots on all Lava environments, comparing \Algname{} ($\lambda$=-1.5) to \DDQN{} and \TempoRL{} on logarithmically decaying $\epsilon$-strategy (Blue represents states visited more often and white states rarely or never seen).}
    \label{fig:grid_coverage_all.pdf}
\end{figure}

\textbf{Coverage.} Figure~\ref{fig:grid_coverage_all.pdf} illustrates state visitation coverage of different agents for 3 different gridworlds environments.
Both \TempoRL{} and \Algname{} repeat the chosen actions, which can lead to a better exploration.
However, \TempoRL{} is not any better than vanilla \DDQN{}, whereas our \Algname{} shows significantly better coverage. 
This implies that a pessimistic \repetition{} policy inducing safe exploration can result in better coverage of the state space.

\textbf{Distribution of \rc{}.} The full version of distributions of \rc{} is presented in Figure~\ref{fig:grid_lava_total_log_skip_hist}.
It shows that the \TempoRL{} selects large \rc{}, close to 7, more often than our \Algname{}. 
We can see \Algname{} maneuver at a smaller scale as the \riskparameter{} decreases to induce more pessimistic behavior. 
The portion of small \rc{}s tends to increase as being more pessimistic.


\begin{table*}[h]
\centering
\begin{tabular}{ll|ccccc|ccc}
\toprule
               &   & \multicolumn{5}{c}{\TEE{} ($\mu$)} & \multicolumn{3}{c}{\Algname{} ($\lambda$)} \\
  Environment  &   & 1.25  & 1.5   & 2.0     & 2.5   & 3.0   & -0.5 & -1.0 & -1.5   \\

\midrule
\multicolumn{1}{l|}{\multirow{3}{*}{Cliff}} & Linear   &  \textbf{0.80}     &  0.79              & 0.79      &  0.78     &  0.77    & 0.89 & 0.88 & \textbf{0.90}  \\
\multicolumn{1}{l|}{}                        & Log     & 0.92 & \textbf{0.93}           & 0.91      & 0.84      & 0.81     &  0.94 &  0.95  & \textbf{0.96}   \\                   
\multicolumn{1}{l|}{}                        & Fixed   &  \textbf{0.65}     &    0.64            &  0.64     &  0.64     &  0.63    & 0.84   &  0.84  & \textbf{0.85}   \\           
\midrule
\multicolumn{1}{l|}{\multirow{3}{*}{Bridge}} & Linear &  0.75     &     0.75           &  \textbf{0.76}     &  0.74     &  0.73    & 0.83 & 0.84 & \textbf{0.86} \\                   
\multicolumn{1}{l|}{}                        & Log    &  \textbf{0.92} & 0.92 & 0.91 & 0.90 & 0.89   & 0.85 & 0.88 & \textbf{0.92} \\                       
\multicolumn{1}{l|}{}                        & Fixed  &  \textbf{0.59}     &     0.57           &  0.58     &  0.55       & 0.55 & 0.72 & 0.82 & \textbf{0.83} \\                          
\midrule
\multicolumn{1}{l|}{\multirow{3}{*}{Zigzag}} & Linear &  0.62     &     \textbf{0.63}           & 0.61       &   0.61    &   0.62     & 0.73 & 0.82 & \textbf{0.84} \\                 
\multicolumn{1}{l|}{}                        & Log    & \textbf{0.76} & 0.72 & 0.63 & 0.61 & 0.52  & 0.66 & 0.86 & \textbf{0.89} \\                         
\multicolumn{1}{l|}{}                        & Fixed  &  0.36     &     0.40           & 0.41       &   0.42    &    \textbf{0.43}    & 0.62 & 0.70 & \textbf{0.76}\\
\bottomrule
\end{tabular}
\caption{
Normalized AUC of reward for varying hyperparameters of $\mu$ in \TEE{} and $\lambda$ in \Algname{} on a logarithmically decaying $\epsilon$-strategy in the Gridworlds environment. (20 random seeds)
}
\label{tab:grid_tee_mu}
\end{table*}
%%%%%%%%%%%%%%%%%%%%%%%%% Atari %%%%%%%%%%%%%%%%%%%%%%%%%%%%%
\clearpage
\subsection{Atari 2600}
\label{app:further_atari}
\textbf{DQN-normalized score.}
The DQN-normalized score is defined as 
\begin{center}
$score = \displaystyle\frac{\mbox{agent - random}}{\mbox{DQN - random}}$
\end{center}
where agent, random, and DQN are the per-game mean rewards over the last 100,000 time steps for the agent, a random policy, and a \DDQN{} respectively.
We used this DQN-normalized score to summarize the results across various games.

The results in Table~\ref{tab:atari_result_normalized} show mean DQN-normalized scores over the last 100,000 time steps of each game. 
The reason that we use this metric instead of the human-normalized score is that we have only trained the agent for 10 million frames due to limited resources. 
The agents were not fully trained to be compared with the human scores, so we normalized the score against \DDQN{}. 
A score below 0 means that the performance is worse than that of random policy while the score greater than 1 indicates it achieves higher performance compared to that of \DDQN{} agent.
Overall, our \Algname{} with the best uncertainty parameter performs best compared to other baselines (see Table~\ref{tab:atari_result_normalized} for more details). 
\Algname{} achieves a score 81\% higher than that of \DDQN{} and 40\% higher than \TempoRL{}. \\

\begin{table*}[ht]
\centering
    \begin{tabular}{l | cccccc | cc} 
     \toprule
    \multirow{2}{*}{Environment} & \multirow{2}{*}{\DDQN} & \multirow{2}{*}{Fixed-$j$} & \multirow{2}{*}{\TEE{}} & \multirow{2}{*}{\DAR} & \multirow{2}{*}{\TempoRL} & \multirow{2}{*}{\BootstrappedDQN{}} & \Algname{} & \Algname{} \\
    &&&&&&&($1$-step)&  ($n$-step)\\
     \midrule
     Beam Rider      & 1.00 & 0.85 & 2.50           & 0.05 & 2.79           & 1.32 & 2.41& \textbf{2.89}  \\
     Centipede       & 1.00 & 0.55 & 0.82           & 0.79 & 1.49           & 0.63 & 0.82& \textbf{1.71}  \\
     Crazy Climber   & 1.00 & 0.70 & 1.01           & 0.38 & 0.93           & 0.55 & 1.29& \textbf{1.56}  \\
     Freeway         & 1.00 & 0.93 & 0.94           & 0.83 & \textbf{1.18}  & 1.17 & 1.17& 1.13  \\
     Kangaroo        & 1.00 & 0.41 & 1.11           & 0.56 & 0.78           & 0.98 & 1.07 &1.21  \\
     Ms Pacman       & 1.00 & 0.71 & \textbf{1.21}  & 0.67 & 0.97           & 1.02 & \textbf{1.33} &1.10  \\ 
     Pong            & 1.00 & 0.02 & \textbf{1.02}  & 0.01 & 0.91           & 0.98 & 1.00 & 1.00 \\
     Qbert           & 1.00 & 0.48 & 1.41           & 0.38 & 1.08           & 1.69 & 1.39& \textbf{2.09}  \\
     Riverraid       & 1.00 & 0.46 & \textbf{1.29}  & 0.09 & 1.10           & 1.15 & 1.29& 1.28 \\
     Road Runner     & 1.00 & 0.38 & 1.14           & 0.26 & 2.48           & 1.52 & 1.51& \textbf{3.76}  \\
     Sea Quest       & 1.00 & 0.14 & 1.04           & 0.12 & 0.58           & 0.67 & 1.00& \textbf{1.54}  \\
     Up n Down       & 1.00 & 1.13 & 1.63           & 0.58 & 1.23           & 0.66 & 1.83& \textbf{2.05}  \\
     \midrule
     Average       & 1.00 & 0.56 & 1.28           & 0.35 & 1.29          & 1.03 & 1.34 & \textbf{1.81}  \\
     \bottomrule
    \end{tabular}
\caption{
DQN-normalized performance averaged over last 100,000 time steps for \Algname{} with the best \riskparameter{} and other baselines. (7 random seeds)
}
\label{tab:atari_result_normalized}
\end{table*}

\textbf{Per-game Best Parameter.} We applied various kinds of uncertainty parameters to our proposed model from +1.0 to -1.5.
As shown in Table~\ref{tab:atari_result_rs}, the optimal uncertainty parameter varies from environment to environment.
In most games, such as \textit{Beam Rider, Centipede, Crazy Climber, Freeway, Qbert, Road Runner, Up n Down}, the uncertainty-averse strategy (negative $\lambda$) exhibits an improvement in averaged rewards over the last 100,000 time steps.  
Meanwhile, on \textit{Kangaroo}, the exploration-favor strategy (positive $\lambda$) shows better performance.
\\
Table~\ref{tab:atari_tee_mu} describes that optimal hyperparameter $\mu$ for \TEE{} also varies from environment to environment. 
For fair comparison with our algorithm, we used the per-game best $\mu$ for \TEE{}.

\textbf{Multi-arm Bandit for Choosing $\lambda$.}
To choose $\lambda$ adaptively, we used multi-armed bandit (MAB) algorithm~\cite{garivier2008upper} with sliding-window upper confidence bound (UCB) as described in Atari 2600 experiments.
Therefore, here we describe the bandit algorithm in detail.
The following method is mainly structured by referring to Appendix Section D in~\citet{badia2020agent57}.

At each episode $k \in [K]$, a $N$-armed bandit selects an arm $A_k$ among the pre-defined set of arms $\mathcal{A}:= \{0, \dots, N-1 \}$ by a policy $\pi$. 
The policy $\pi$ depends on the sequence of previous histories (actions and rewards).
Then, it receives a reward $R_k(A_k) \in \mathbb{R}$ from the environment.

The objective of an MAB algorithm is to learn a policy $\pi$ that minimizes the expected regret as follows:
\begin{align*}
    \mathbb{E}_{\pi} \left[ \sum_{k=0}^{K-1} \max_A R_k(A) -   R_k(A_k) \right].
\end{align*}

When reward distribution is stationary, i.e. $R_k(\cdot) = R(\cdot)$, the traditional UCB algorithm can be applied.
Define the number of time episodes an arm $a \in \mathcal{A}$ has been selected in episode $k$ as:
\begin{align*}
    N_k(a) = \sum_{k'=0}^{k-1} \bm{1}(A_{k'} = a),
\end{align*}
where $\bm{1}(A_k = a)$ is an indicator function. 
We can estimate the empirical mean reward of an arm $a$ as:
\begin{align*}
    \widehat{\mu}_k(a) = \frac{1}{N_k(a)}\sum_{k'=0}^{k=1}R_{k'}(a) \bm{1}(A_{k'} = a).
\end{align*}
Then, we select an arm using the UCB algorithm as follows:
\begin{align*}
    \begin{cases}
        \forall 0 \leq k \leq N-1, & A_k = k, \\
        \forall N \leq k \leq K-1, & A_k = \arg\!\max_{a \in \mathcal{A}} \widehat{\mu}_{k-1}(a) + \beta \sqrt{\frac{\log(k-1)}{N_{k-1}(a)}}.
    \end{cases}
\end{align*}
However, if reward distribution is non-stationary, the UCB algorithm cannot be directly applied due to the change in reward distribution.
One of the common solutions to the non-stationary case is to use a sliding-window UCB.
Let $\tau \in \mathbb{Z}^{+}$ be the size of window such that $\tau < K$.
The number of time episodes an arm $a \in \mathcal{A}$ has been played in episode $k$ for a window size $\tau$ as:
\begin{align}\label{eq:number_sliding}
    N_k^{\tau}(a) = \sum_{k'= \max(0, k-\tau) }^{k-1} \bm{1}(A_{k'} = a).
\end{align}
Define the empirical mean reward of an arm $a$ for a window size $\tau$ as:
\begin{align}\label{eq:empirical_mean_reward_sliding}
    \widehat{\mu}_k^{\tau}(a) = \frac{1}{N_k^{\tau}(a)}\sum_{k'= \max(0, k-\tau)}^{k=1}R_{k'}(a) \bm{1}(A_{k'} = a).
\end{align}
Then, we select an arm using the sliding window UCB as follows:
\begin{align*}
    \begin{cases}
        \forall 0 \leq k \leq N-1, & A_k = k, \\
        \forall N \leq k \leq K-1 & A_k = \arg\!\max_{a \in \mathcal{A}} \widehat{\mu}_{k-1}^{\tau}(a) + \beta \sqrt{\frac{\log(k-1)}{N_{k-1}^{\tau}(a)}}.
    \end{cases}
\end{align*}
Finally, since we use the sliding window UCB with $\epsilon_{ucb}$-greedy exploration, our bandit algorithm is as follows:
\begin{align*}
    \begin{cases}
        \forall 0 \leq k \leq N-1, & A_k = k, \\
        \forall N \leq k \leq K-1 \,\,\text{and}\,\, U_k \geq \epsilon_{ucb},   & A_k = \arg\!\max_{a \in \mathcal{A}} \widehat{\mu}_{k-1}^{\tau}(a) + \beta \sqrt{\frac{\log(k-1)}{N_{k-1}^{\tau}(a)}}, \\
        \forall N \leq k \leq K-1 \,\,\text{and}\,\, U_k < \epsilon_{ucb},   & A_k = Y_k,
    \end{cases}
\end{align*}
where $U_k$ is a random variable drawn uniformly from $[0,1]$ and $Y_k$ is a random action sampled uniformly from $\mathcal{A} = \{ 0, \dots, N-1 \}$.

\begin{figure}[ht]
    \centering
     \includegraphics[clip, trim=4.7cm 0.0cm 4.3cm 0.2cm, width=1.0\textwidth]{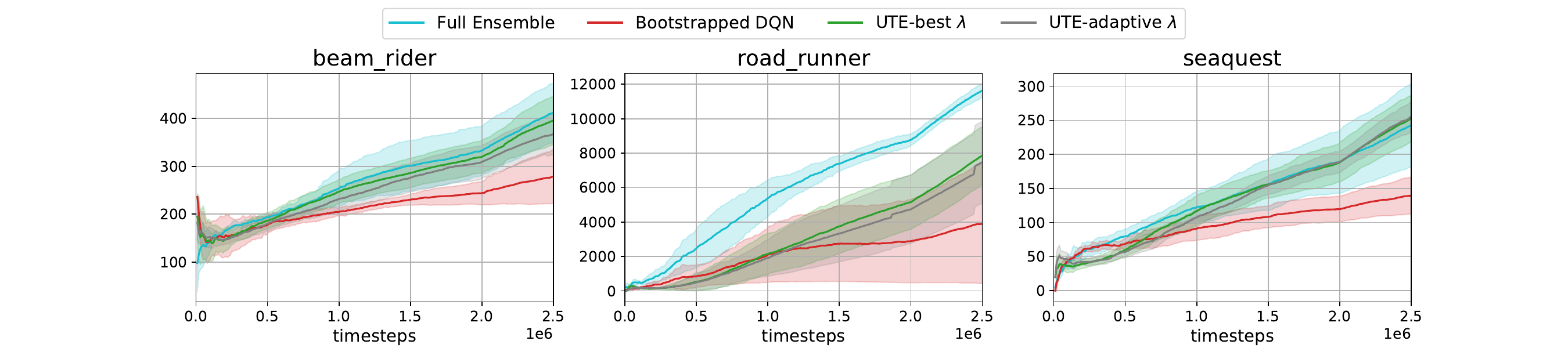}
     \caption{Learning curves of \Full{}, \BootstrappedDQN{}, \Algname{} with the best uncertainty parameter, and \Algname{} with adaptive uncertainty parameter over 7 random seeds.}
     \label{fig:atari_full_learning_curve} 
\end{figure}

In Atari experiments, each arm corresponds to \rc{} $\lambda$.
At the beginning of each episode, the bandit algorithm chooses $\lambda_k$ among the set, $\lambda_k \in \Lambda := \{ +1.0, +0.5, +0.2, 0.0, -0.2, -0.5, -1.0, -1.5 \}$, and gets the feedback of episode rewards $R_k(\lambda_k)$.
Then, the bandit algorithm update $N_k^{\tau}(a)$ by Eq.~\eqref{eq:number_sliding} and $ \widehat{\mu}_k^{\tau}(a)$ by Eq.~\eqref{eq:empirical_mean_reward_sliding}.

\textbf{Full Ensemble Model.}
We additionally evaluated another algorithm, called \Full{}. 
The \Full{} is a combination of \Algname{} and \BootstrappedDQN{}, which means both action-value functions and option-value functions are estimated by an ensemble method. 
Figure~\ref{fig:atari_full_learning_curve} demonstrates that the \Full{} performs similar to or better than others.
In an environment where rewards are relatively sparse such as \textit{Road Runner}, \Full{} notably outperforms other agents.
We did not optimize the uncertainty parameter for \Full{} so that there is room for further improvements.
The results imply that our method can apply to any base algorithm smoothly.\\

\begin{figure}[ht]
    \centering
        \includegraphics[clip, trim=0.0cm 2.0cm 11.0cm 2.0cm, width=0.45\textwidth]{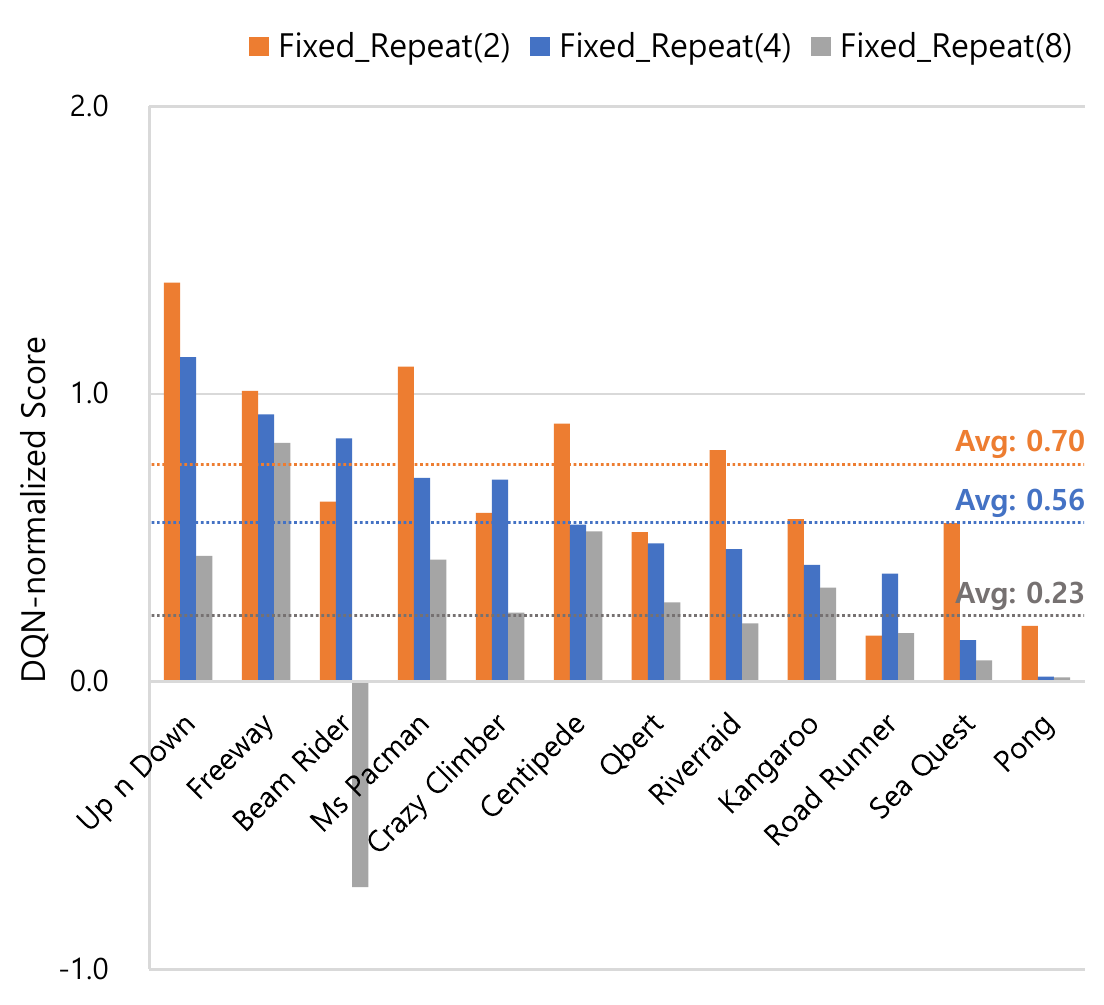}  %trim = left, botm, right, top
    \caption{DQN-normalized score for Fixed Repeat with varying fixed $j$. (5 random seeds)}
    \label{fig:appendix_fixed_j}
\end{figure}
\begin{figure}[ht]
    \centering
     \begin{subfigure}[b]{0.4\textwidth}
         \centering
         \includegraphics[clip, trim=0.0cm 3.0cm 11.0cm 2.0cm, width=1.00\textwidth]{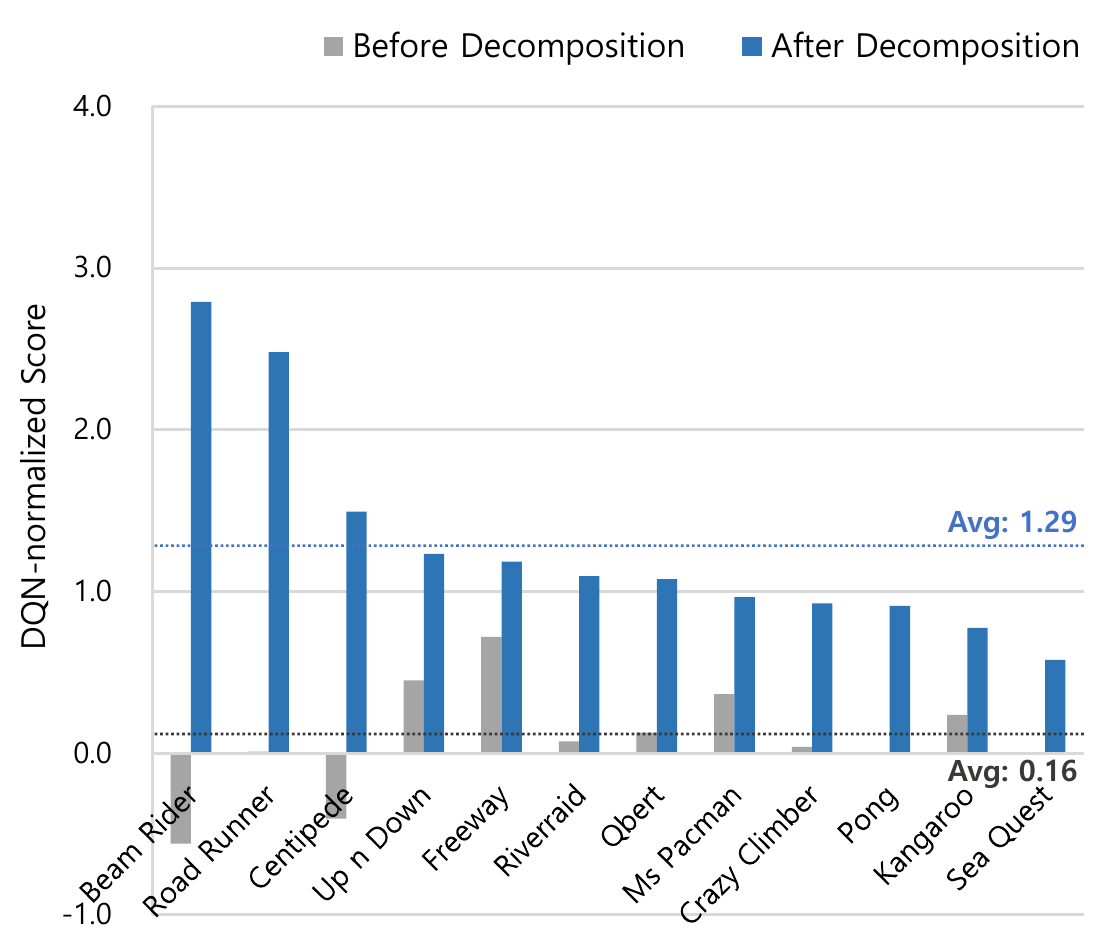}
         \caption{Decomposition Effect}
         \label{fig:ablation_decomposition}
     \end{subfigure}
     \hspace*{1cm}
     \begin{subfigure}[b]{0.4\textwidth}
         \centering
         \includegraphics[clip, trim=0.0cm 3.0cm 11.0cm 2.0cm, width=1.00\textwidth]{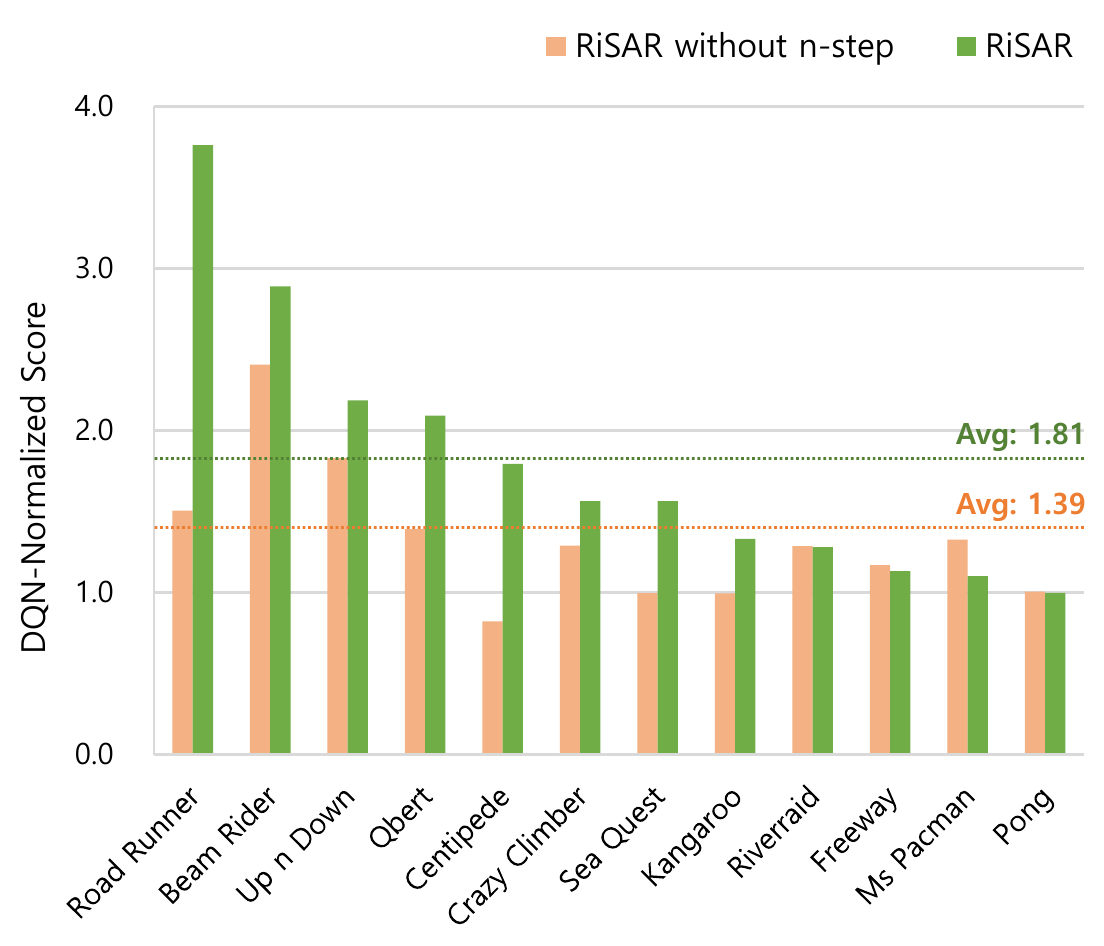}
         \caption{Multi-step Target Effect for $\behaviorQ{}$}
        \label{fig:ablation_nstep}
     \end{subfigure}
 \caption{Ablation studies for decomposition effect (left) and $n$-step learning effect for action-value function $\behaviorQ{}$. The negative score indicates that the policy is worse than a random policy. (5 random seeds)}
\label{fig:ablations}
\end{figure}

\textbf{Various Fixed Repeat.}
Figure~\ref{fig:appendix_fixed_j} shows the performance of Fixed Repeat agents with varying fixed \rc{}. 
The fixed \rc{} affects the granularity of control. 
The negative score indicates that the performance is worse than a random policy.
The result describes that naively repeating the chosen action could degrade the performance in Atari environments.
And this tends to get worse as the fixed \rc{} is increased.
However, it still shows a lot worse performance than \DQN{} (score = $1.0$).

\textbf{Ablation 1: Decomposition.}
Our method formulates joint optimization of the action and the \rc{} as a two-level optimization problem. 
The action is selected based on $\behaviorQ{}$ and then the \rc{} is selected based on $\skipQ{}$ sequentially. 
Left of Figure~\ref{fig:ablation_decomposition} shows the effect of the decomposition. 
Without decomposition, the size of search space is $|\mathcal{A}| \times |J|$, which leads to a catastrophic performance.
In some environments such as \textit{Beam Rider} and \textit{Road Runner}, the DQN-normalized scores are negative, which means the agent is worse than a random policy.
Overall, We can see that decomposition of action and \rc{} selection improves the performance significantly.

\textbf{Ablation 2: Multi-step Target.}
Right of Figure~\ref{fig:ablation_nstep} describes the effect of using an n-step target for $\behaviorQ{}$. 
We compare \Algname{} with n-step Q-learning to the one without it. 
This result illustrates that applying $n$-step learning is beneficial in most games, which shows a $30.2\%$ improvement (from 1.39 to 1.81) after it has been applied.
Especially in games with relatively sparse rewards such as \textit{Road Runner} and \textit{Centipede}, it dramatically enhanced the performance.
This is because rewards can be propagated faster using $n$-step returns. 

%%%%%%%%%%%%%%%%%%%%%%%%%%%%%%%%%%%%%%%%%%%%%%%%%%%%
\begin{table*}[hbt!]
\centering
    \begin{tabular}{l|cccccc}
     \toprule
                     & \multicolumn{5}{c}{\TEE{} ($\mu$)} \\
      Environment     & 1.5  & 1.75   & 2.0     & 2.25   & 2.5     \\
     \midrule
     Beam Rider      & 331.6 & 272.9 & \textbf{409.1}           & 261.4 & 328.9            \\
     Centipede       & 1271.8 & 1222.6 & 1080.2          & 1316.0 & \textbf{1431.7}           \\
     Crazy Climber   & 5026.1 & 4420.0 & 3128.0           & \textbf{5295.1} &4690.9             \\
     Freeway         & \textbf{30.8} & 30.7 & 25.6           & 20.5 & 25.6   \\
     Kangaroo        & \textbf{609.1} & 518.8 & 604.0           & 360.0 & 396.4             \\
     Ms Pacman       & 580.0 & 551.3 & 584.5  & 514.9 & \textbf{597.2}            \\ 
     Pong            & 19.7 & 18.4 & 19.8  & 19.4 & \textbf{19.9}           \\
     Qbert           & \textbf{392.8} & 388.6 & 345.2          & 264.7 & 270.6            \\
     Riverraid       & 810.4 & 835.5 & \textbf{945.6}  & 695.5 & 738.2            \\
     Road Runner     & 2943.0 & 2215.2 & \textbf{3733.8}           & 3131.2 & 848.5             \\
     Sea Quest       & 116.1 & \textbf{214.5} & 172.6           & 123.8 & 119.6             \\
     Up n Down       & 700.6 & \textbf{823.1} & 669.0          & 794.8 & 653.2            \\
     \bottomrule
    \end{tabular}
\caption{
Average rewards for \TEE{} varying values for hyperparameter $\mu$ in the Atari 2600 environments. (7 random seeds)
}
\label{tab:atari_tee_mu}
\end{table*}
%%
%%%%%%%%%%%%%%%%%%%%%%%%%%%%%%%%%%%%%%%%%%
\begin{table*}[hbt!]
\centering
    \begin{tabular}{l | rrrrrrrr} 
    \toprule
     \ & \multicolumn{8}{c}{\Algname{} (\riskparameter{}: $\lambda$)} \\
     Environment & +1.0 & +0.5 & +0.2 & +0.0 & -0.2 & -0.5 & -1.0 & -1.5\\
     \midrule
     Beam Rider & \begin{tabular}[r]{@{}r@{}}384.3\\ {\small (137.8)}\end{tabular} & 
                \begin{tabular}[r]{@{}r@{}}431.6\\ {\small (162.9)}\end{tabular} &
                \begin{tabular}[r]{@{}r@{}}401.1\\ {\small (143.3)}\end{tabular} & 
                \begin{tabular}[r]{@{}r@{}}425.2\\ {\small (153.5)}\end{tabular} & 
                \begin{tabular}[r]{@{}r@{}}403.1\\ {\small (127.9)}\end{tabular} &
                \begin{tabular}[r]{@{}r@{}}417.1\\ {\small (115.4)}\end{tabular} &
                \begin{tabular}[r]{@{}r@{}}\textbf{439.5}\\ {\small (163.0)}\end{tabular} &
                \begin{tabular}[r]{@{}r@{}}414.2 \\ {\small (112.3)}\end{tabular}  \\
     
     Centipede & \begin{tabular}[r]{@{}r@{}}1898.2\\ {\small (889.2)}\end{tabular} & 
                \begin{tabular}[r]{@{}r@{}}1327.8 \\ {\small (760.6)}\end{tabular} &
                \begin{tabular}[r]{@{}r@{}}1581.4\\ {\small (1008.2)}\end{tabular} & 
                \begin{tabular}[r]{@{}r@{}}2125.6\\ {\small (1356.4)}\end{tabular} &
                \begin{tabular}[r]{@{}r@{}}\textbf{2190.1}\\ {\small (1073.0)}\end{tabular}  & 
                \begin{tabular}[r]{@{}r@{}}1893.6\\ {\small (954.9)}\end{tabular} &
                \begin{tabular}[r]{@{}r@{}}1605.9\\ {\small (1167.7)}\end{tabular} & 
                \begin{tabular}[r]{@{}r@{}}1377.5\\ {\small (875.2)}\end{tabular}   \\
     
     Crazy Climber & \begin{tabular}[r]{@{}r@{}}5093.2\\ {\small (4011.6)}\end{tabular} & 
                \begin{tabular}[r]{@{}r@{}}4426.2\\ {\small (3987.4)}\end{tabular} & 
                \begin{tabular}[r]{@{}r@{}}6163.6\\ {\small (5025.2)}\end{tabular} & 
                \begin{tabular}[r]{@{}r@{}}5033.9\\ {\small (2653.9)}\end{tabular} &
                \begin{tabular}[r]{@{}r@{}}6484.8\\ {\small (4797.2)}\end{tabular} & 
                \begin{tabular}[r]{@{}r@{}}\textbf{8175.6}\\ {\small (5790.4)}\end{tabular} &
                \begin{tabular}[r]{@{}r@{}}5198.6\\ {\small (4049.5)}\end{tabular} & 
                \begin{tabular}[r]{@{}r@{}}5220.2\\ {\small (4751.1)}\end{tabular}   \\
     
     Freeway & \begin{tabular}[r]{@{}r@{}}28.6\\ {\small (3.7)}\end{tabular} & 
                \begin{tabular}[r]{@{}r@{}}29.9\\ {\small (2.0)}\end{tabular} & 
                \begin{tabular}[r]{@{}r@{}}27.9\\ {\small (5.4)}\end{tabular} &
                \begin{tabular}[r]{@{}r@{}}30.5\\ {\small (1.5)}\end{tabular} & 
                \begin{tabular}[r]{@{}r@{}}\textbf{30.7}\\ {\small (1.9)}\end{tabular} & 
                \begin{tabular}[r]{@{}r@{}}24.1\\ {\small (3.9)}\end{tabular} &
                \begin{tabular}[r]{@{}r@{}}25.1\\ {\small (4.7)}\end{tabular} & 
                \begin{tabular}[r]{@{}r@{}}25.0\\ {\small (6.8)}\end{tabular}   \\
     
     Kangaroo  & \begin{tabular}[r]{@{}r@{}}555.2\\ {\small (650.7)}\end{tabular} & 
                \begin{tabular}[r]{@{}r@{}}493.3\\ {\small (515.7)}\end{tabular} & 
                \begin{tabular}[r]{@{}r@{}}543.0\\ {\small (450.3)}\end{tabular} &
                \begin{tabular}[r]{@{}r@{}}661.0\\ {\small (630.4)}\end{tabular} &
                \begin{tabular}[r]{@{}r@{}}623.3\\ {\small (630.6)}\end{tabular} & 
                \begin{tabular}[r]{@{}r@{}}577.6\\ {\small (622.6)}\end{tabular} &
                \begin{tabular}[r]{@{}r@{}}718.2\\ {\small (859.1)}\end{tabular} & 
                \begin{tabular}[r]{@{}r@{}}\textbf{728.5}\\ {\small (832.6)}\end{tabular}   \\
     
     Ms Pacman  & \begin{tabular}[r]{@{}r@{}}446.5\\ {\small (229.7)}\end{tabular} & 
                \begin{tabular}[r]{@{}r@{}}509.6\\ {\small (256.3)}\end{tabular} & 
                \begin{tabular}[r]{@{}r@{}}551.0\\ {\small (237.5)}\end{tabular} &
                \begin{tabular}[r]{@{}r@{}}545.5\\ {\small (218.3)}\end{tabular} &
                \begin{tabular}[r]{@{}r@{}}\textbf{551.6}\\ {\small (225.5)}\end{tabular} &
                \begin{tabular}[r]{@{}r@{}}544.1\\ {\small (249.1)}\end{tabular} &
                \begin{tabular}[r]{@{}r@{}}511.2\\ {\small (182.2)}\end{tabular} & 
                \begin{tabular}[r]{@{}r@{}}540.4\\ {\small (228.3)}\end{tabular}   \\ 
     
     Pong & \begin{tabular}[r]{@{}r@{}}17.5\\ {\small (5.9)}\end{tabular} & 
                \begin{tabular}[r]{@{}r@{}}18.0\\ {\small (5.1)}\end{tabular} & 
                \begin{tabular}[r]{@{}r@{}}17.1\\ {\small (5.9)}\end{tabular} &
                \begin{tabular}[r]{@{}r@{}}16.9\\ {\small (4.5)}\end{tabular} &
                \begin{tabular}[r]{@{}r@{}}\textbf{19.1}\\ {\small (2.5)}\end{tabular} &
                \begin{tabular}[r]{@{}r@{}}16.8\\ {\small (6.2)}\end{tabular} & 
                \begin{tabular}[r]{@{}r@{}}18.4\\ {\small (4.7)}\end{tabular} & 
                \begin{tabular}[r]{@{}r@{}}16.4\\ {\small (4.6)}\end{tabular}   \\
     
     Qbert  & \begin{tabular}[r]{@{}r@{}}387.4\\ {\small (415.8)}\end{tabular} & 
                \begin{tabular}[r]{@{}r@{}}400.5\\ {\small (490.3)}\end{tabular} & 
                \begin{tabular}[r]{@{}r@{}}457.4\\ {\small (461.3)}\end{tabular} &
                \begin{tabular}[r]{@{}r@{}}399.2\\ {\small (461.9)}\end{tabular} &
                \begin{tabular}[r]{@{}r@{}}297.2 \\ {\small (302.7)}\end{tabular}&
                \begin{tabular}[r]{@{}r@{}}417.1\\ {\small (441.0)}\end{tabular} & 
                \begin{tabular}[r]{@{}r@{}}\textbf{582.5}\\ {\small (558.7)}\end{tabular} & 
                \begin{tabular}[r]{@{}r@{}}459.1\\ {\small (499.7)}\end{tabular}   \\
     
     Riverraid & \begin{tabular}[r]{@{}r@{}}\textbf{938.0}\\ {\small (388.8)}\end{tabular} & 
                \begin{tabular}[r]{@{}r@{}}828.6\\ {\small (339.6)}\end{tabular} & 
                \begin{tabular}[r]{@{}r@{}}823.3\\ {\small (363.0)}\end{tabular} &
                \begin{tabular}[r]{@{}r@{}}922.1\\ {\small (418.0)}\end{tabular} &
                \begin{tabular}[r]{@{}r@{}}880.3\\ {\small (341.0)}\end{tabular} & 
                \begin{tabular}[r]{@{}r@{}}909.8\\ {\small (350.1)}\end{tabular} & 
                \begin{tabular}[r]{@{}r@{}}863.1\\ {\small (354.8)}\end{tabular} & 
                \begin{tabular}[r]{@{}r@{}}795.8\\ {\small (310.5)}\end{tabular}   \\
     
     Road Runner & \begin{tabular}[r]{@{}r@{}}10051.5\\ {\small (4107.8)}\end{tabular} & 
                \begin{tabular}[r]{@{}r@{}}10853.3\\ {\small (5789.4)}\end{tabular} & 
                \begin{tabular}[r]{@{}r@{}}10712.7\\ {\small (3290.1)}\end{tabular} &
                \begin{tabular}[r]{@{}r@{}}9788.2\\ {\small (4054.3)}\end{tabular} &
                \begin{tabular}[r]{@{}r@{}}9019.5\\ {\small (4469.1)}\end{tabular} & 
                \begin{tabular}[r]{@{}r@{}}\textbf{12323.2}\\ {\small (4177.1)}\end{tabular} &
                \begin{tabular}[r]{@{}r@{}}6638.6\\ {\small (3602.0)}\end{tabular} & 
                \begin{tabular}[r]{@{}r@{}}5763.6 \\ {\small (4681.0)}\end{tabular}  \\
     
     Sea Quest & \begin{tabular}[r]{@{}r@{}}260.3\\ {\small (126.7)}\end{tabular} & 
                \begin{tabular}[r]{@{}r@{}}301.0\\ {\small (162.6)}\end{tabular} & 
                \begin{tabular}[r]{@{}r@{}}290.4\\ {\small (154.4)}\end{tabular} &
                \begin{tabular}[r]{@{}r@{}}308.5\\ {\small (150.8)}\end{tabular} &
                \begin{tabular}[r]{@{}r@{}}\textbf{313.4}\\ {\small (141.1)}\end{tabular} &
                \begin{tabular}[r]{@{}r@{}}282.4\\ {\small (164.7)}\end{tabular} & 
                \begin{tabular}[r]{@{}r@{}}250.9 \\ {\small (162.0)}\end{tabular}& 
                \begin{tabular}[r]{@{}r@{}}226.8 \\ {\small (125.4)}\end{tabular}  \\
     
     Up n Down  & \begin{tabular}[r]{@{}r@{}}1012.7\\ {\small (613.6)}\end{tabular} & 
                \begin{tabular}[r]{@{}r@{}}999.6\\ {\small (504.3)}\end{tabular} & 
                \begin{tabular}[r]{@{}r@{}}865.8\\ {\small (451.2)}\end{tabular} &
                \begin{tabular}[r]{@{}r@{}}912.1\\ {\small (611.0)}\end{tabular} &
                \begin{tabular}[r]{@{}r@{}}990.3\\ {\small (532.0)}\end{tabular} & 
                \begin{tabular}[r]{@{}r@{}}972.5\\ {\small (530.2)}\end{tabular} & 
                \begin{tabular}[r]{@{}r@{}}\textbf{1072.8}\\ {\small (664.0)}\end{tabular} & 
                \begin{tabular}[r]{@{}r@{}}1039.6 \\ {\small (573.0)}\end{tabular}  \\
     \bottomrule
    \end{tabular}
\caption{
Average rewards and standard deviations (numbers in bracket) over the last 100,000 time steps for different \riskparameter{} of our proposed method. (7 random seeds)
}
\label{tab:atari_result_rs}
\end{table*}
%%%%%%%%%%%%%%%%%%%%%%%%%%%%%%%%%%%%%%%%%%
\begin{table*}[]
\centering 
\resizebox{\linewidth}{!}{
\begin{tabular}{l | rrrrrr | rrr} 
\toprule
% & \multicolumn{7}{c}{B}      \\
\multirow{2}{*}{Environment} & \multirow{2}{*}{\DDQN} & \multirow{2}{*}{Fixed-$j$} & \multirow{2}{*}{\TEE{}} & \multirow{2}{*}{\DAR} & \multirow{2}{*}{\TempoRL} & \multirow{2}{*}{\BootstrappedDQN} & \multicolumn{3}{c}{\Algname{}}  \\
&&&&&&&$1$-step&  $n$-step & Adaptive $\lambda$\\
\midrule
 Beam Rider & \begin{tabular}[r]{@{}r@{}}290.1\\ {\small (101.5)}\end{tabular}& 
                \begin{tabular}[r]{@{}r@{}}277.9\\ {\small (109.9)}\end{tabular} & 
                \begin{tabular}[r]{@{}r@{}}409.9\\ {\small (124.0)}\end{tabular} & 
                \begin{tabular}[r]{@{}r@{}}177.7\\ {\small (73.3)}\end{tabular}  & 
                \begin{tabular}[r]{@{}r@{}}431.9\\ {\small (140.4)}\end{tabular}& 
                \begin{tabular}[r]{@{}r@{}}315.6\\ {\small (131.1)}\end{tabular} & 
                
                \begin{tabular}[r]{@{}r@{}}414.4\\ {\small (141.1)}\end{tabular} & 
                \begin{tabular}[r]{@{}r@{}}\textbf{439.5}\\ {\small (163.0)}\end{tabular} &
                \begin{tabular}[r]{@{}r@{}}423.9\\ {\small (158.5)}\end{tabular}\\
                
 Centipede  &  \begin{tabular}[r]{@{}r@{}}1574.7\\{\small (1044.6)} \end{tabular}& 
            \begin{tabular}[r]{@{}r@{}}1222.8\\ {\small (840.8)}\end{tabular}  & 
            \begin{tabular}[r]{@{}r@{}}1431.7\\ {\small (1169.1)}\end{tabular} & 
            \begin{tabular}[r]{@{}r@{}}1410.3\\ {\small (982.8)}\end{tabular}  & 
            \begin{tabular}[r]{@{}r@{}}1958.0\\ {\small (1166.4)}\end{tabular}  & 
            \begin{tabular}[r]{@{}r@{}}1285.2\\ {\small (867.2)}\end{tabular} & 
            
            \begin{tabular}[r]{@{}r@{}}1437.1\\ {\small (887.2)}\end{tabular} & 
            \begin{tabular}[r]{@{}r@{}}\textbf{2190.1}\\ {\small (1073.0)}\end{tabular} &
            \begin{tabular}[r]{@{}r@{}}1829.9\\ {\small (969.6)}\end{tabular}\\
            
 Crazy Climber  & \begin{tabular}[r]{@{}r@{}}5265.8\\{\small (4063.4)} \end{tabular} & 
            \begin{tabular}[r]{@{}r@{}}3731.1\\ {\small (2997.2)}\end{tabular} & 
            \begin{tabular}[r]{@{}r@{}}5295.1\\ {\small (3609.7)}\end{tabular} & 
            \begin{tabular}[r]{@{}r@{}}2059.1\\ {\small (1225.1)}\end{tabular}  & 
            \begin{tabular}[r]{@{}r@{}}4885.5\\ {\small (3378.3)}\end{tabular}  & 
            \begin{tabular}[r]{@{}r@{}}2961.6\\ {\small (3080.0)}\end{tabular} & 
            
            \begin{tabular}[r]{@{}r@{}}6761.9\\ {\small (5061.9)}\end{tabular} & 
            \begin{tabular}[r]{@{}r@{}}\textbf{8175.6}\\ {\small (5790.4)}\end{tabular} &
            \begin{tabular}[r]{@{}r@{}}7046.3\\ {\small (5350.0)}\end{tabular} \\
            
 Freeway  & \begin{tabular}[r]{@{}r@{}}27.1\\{\small (4.6)} \end{tabular} & 
            \begin{tabular}[r]{@{}r@{}}25.2\\ {\small (3.1)}\end{tabular}  & 
            \begin{tabular}[r]{@{}r@{}}30.8\\ {\small (1.3)}\end{tabular}  & 
            \begin{tabular}[r]{@{}r@{}}22.5\\ {\small (2.9)}\end{tabular} & 
            \begin{tabular}[r]{@{}r@{}}\textbf{32.1}\\ {\small (1.0)}\end{tabular}  & 
            \begin{tabular}[r]{@{}r@{}}31.7\\ {\small (1.1)}\end{tabular} & 
            
            \begin{tabular}[r]{@{}r@{}}31.7\\ {\small (1.2)}\end{tabular} & 
            \begin{tabular}[r]{@{}r@{}}30.7\\ {\small (1.9)}\end{tabular} &  
            \begin{tabular}[r]{@{}r@{}}30.8\\ {\small (1.7)}\end{tabular}\\
            
 Kangaroo  & \begin{tabular}[r]{@{}r@{}}547.2\\{\small (764.9)} \end{tabular} & 
            \begin{tabular}[r]{@{}r@{}}222.2\\ {\small (247.9)}\end{tabular}  & 
            \begin{tabular}[r]{@{}r@{}}609.1\\ {\small (880.1)}\end{tabular}  & 
            \begin{tabular}[r]{@{}r@{}}305.5\\ {\small (340.6)}\end{tabular} & 
            \begin{tabular}[r]{@{}r@{}}424.2\\ {\small (282.0)}\end{tabular}  &
            \begin{tabular}[r]{@{}r@{}}534.8\\ {\small (467.4)}\end{tabular} & 
            
            \begin{tabular}[r]{@{}r@{}}586.4\\ {\small (753.5)}\end{tabular} & 
            \begin{tabular}[r]{@{}r@{}}\textbf{728.5}\\ {\small (832.6)}\end{tabular}  &
            \begin{tabular}[r]{@{}r@{}}661.0\\ {\small (613.5)}\end{tabular}\\
            
 Ms Pacman  & \begin{tabular}[r]{@{}r@{}}509.8\\{\small (204.6)} \end{tabular} & 
            \begin{tabular}[r]{@{}r@{}}388.8\\ {\small (201.6)}\end{tabular}  & 
            \begin{tabular}[r]{@{}r@{}}597.2\\ {\small (261.5)}\end{tabular}  &
            \begin{tabular}[r]{@{}r@{}}371.3\\ {\small (166.5)}\end{tabular} & 
            \begin{tabular}[r]{@{}r@{}}495.6\\ {\small (245.1)}\end{tabular}  & 
            \begin{tabular}[r]{@{}r@{}}516.1\\ {\small (206.5)}\end{tabular} & 
            
            \begin{tabular}[r]{@{}r@{}}\textbf{645.6}\\ {\small (267.1)}\end{tabular} & 
            \begin{tabular}[r]{@{}r@{}}551.6 \\ {\small (225.5)}\end{tabular} &
            \begin{tabular}[r]{@{}r@{}}537.6\\ {\small (263.8)}\end{tabular}\\ 
            
 Pong  & \begin{tabular}[r]{@{}r@{}}19.2\\{\small (4.8)} \end{tabular} & 
            \begin{tabular}[r]{@{}r@{}}-20.3\\ {\small (0.9)}\end{tabular}  & 
            \begin{tabular}[r]{@{}r@{}}\textbf{19.9}\\ {\small (1.8)}\end{tabular}  & 
            \begin{tabular}[r]{@{}r@{}}-20.6\\ {\small (0.6)}\end{tabular} & 
            \begin{tabular}[r]{@{}r@{}}15.6\\ {\small (7.7)}\end{tabular}  & 
            \begin{tabular}[r]{@{}r@{}}18.3\\ {\small (3.6)}\end{tabular} & 
            
            \begin{tabular}[r]{@{}r@{}}19.4\\ {\small (1.8)}\end{tabular} & 
            \begin{tabular}[r]{@{}r@{}}19.1 \\ {\small (2.5)}\end{tabular} &
            \begin{tabular}[r]{@{}r@{}}19.5\\ {\small (2.3)}\end{tabular}\\
            
 Qbert  & \begin{tabular}[r]{@{}r@{}}278.3\\{\small (312.9)} \end{tabular} & 
            \begin{tabular}[r]{@{}r@{}}133.8 \\ {\small (267.9)}\end{tabular} & 
            \begin{tabular}[r]{@{}r@{}}392.8 \\ {\small (438.3)}\end{tabular} & 
            \begin{tabular}[r]{@{}r@{}}104.7\\ {\small (70.8)}\end{tabular} & 
            \begin{tabular}[r]{@{}r@{}}299.7\\ {\small (349.8)}\end{tabular}  & 
            \begin{tabular}[r]{@{}r@{}}470.8\\ {\small (489.5)}\end{tabular} & 
            
            \begin{tabular}[r]{@{}r@{}}387.3\\ {\small (423.1)}\end{tabular} & 
            \begin{tabular}[r]{@{}r@{}}\textbf{582.5}\\ {\small (558.7)}\end{tabular}  &
            \begin{tabular}[r]{@{}r@{}}581.4\\ {\small (602.5)}\end{tabular}\\
            
 Riverraid  & \begin{tabular}[r]{@{}r@{}}740.5\\{\small (291.9)} \end{tabular}  & 
            \begin{tabular}[r]{@{}r@{}}360.9\\ {\small (231.6)}\end{tabular} & 
            \begin{tabular}[r]{@{}r@{}}\textbf{945.6}\\ {\small (457.9)}\end{tabular}  & 
            \begin{tabular}[r]{@{}r@{}}102.8\\ {\small (66.0)}\end{tabular} & 
            \begin{tabular}[r]{@{}r@{}}807.7\\ {\small (354.7)}\end{tabular}  & 
            \begin{tabular}[r]{@{}r@{}}843.8\\ {\small (407.2)}\end{tabular} &
            
            \begin{tabular}[r]{@{}r@{}}942.2\\ {\small (319.3)}\end{tabular} & 
            \begin{tabular}[r]{@{}r@{}}938.0\\ {\small (388.8)}\end{tabular} &
            \begin{tabular}[r]{@{}r@{}}890.5\\ {\small (330.7)}\end{tabular}
            \\
 Road Runner  & \begin{tabular}[r]{@{}r@{}}3277.0\\{\small (4470.3)} \end{tabular} & 
            \begin{tabular}[r]{@{}r@{}}1230.3\\ {\small (1640.9)}\end{tabular}  & 
            \begin{tabular}[r]{@{}r@{}}3733.8\\ {\small (4716.5)}\end{tabular}  & 
            \begin{tabular}[r]{@{}r@{}}845.5\\ {\small (791.9)}\end{tabular} & 
            \begin{tabular}[r]{@{}r@{}}8131.5\\ {\small (4099.3)}\end{tabular}  &
            \begin{tabular}[r]{@{}r@{}}4976.8 \\ {\small (6032.6)}\end{tabular}& 
            
            \begin{tabular}[r]{@{}r@{}}4935.6\\ {\small (5206.4)}\end{tabular} & 
            \begin{tabular}[r]{@{}r@{}}\textbf{12323.2}\\ {\small (4177.1)}\end{tabular}  &
            \begin{tabular}[r]{@{}r@{}}10353.3\\ {\small (3283.3)}\end{tabular}\\
            
 Sea Quest  & \begin{tabular}[r]{@{}r@{}}207.6\\{\small (124.5)} \end{tabular} & 
            \begin{tabular}[r]{@{}r@{}}47.0\\ {\small (26.2)}\end{tabular} & 
            \begin{tabular}[r]{@{}r@{}}214.5\\ {\small (85.6)}\end{tabular}  & 
            \begin{tabular}[r]{@{}r@{}}42.8\\ {\small (33.9)}\end{tabular} & 
            \begin{tabular}[r]{@{}r@{}}128.2 \\ {\small (55.5)}\end{tabular} & 
            \begin{tabular}[r]{@{}r@{}}145.1\\ {\small (64.5)}\end{tabular} & 
            
            \begin{tabular}[r]{@{}r@{}}206.9\\ {\small (92.4)}\end{tabular} & 
            \begin{tabular}[r]{@{}r@{}}313.4\\ {\small (141.1)}\end{tabular} &
            \begin{tabular}[r]{@{}r@{}}\textbf{320.3}\\ {\small (159.4)}\end{tabular}\\
            
 Up n Down  &\begin{tabular}[r]{@{}r@{}}536.4\\{\small (361.5)} \end{tabular}  & 
            \begin{tabular}[r]{@{}r@{}}594.8\\ {\small (324.6)}\end{tabular}  & 
            \begin{tabular}[r]{@{}r@{}}823.1\\ {\small (320.0)}\end{tabular} & 
            \begin{tabular}[r]{@{}r@{}}348.7\\ {\small (227.0)}\end{tabular} & 
            \begin{tabular}[r]{@{}r@{}}641.5\\ {\small (428.6)}\end{tabular}& 
            \begin{tabular}[r]{@{}r@{}}383.2\\ {\small (242.8)}\end{tabular} & 
            
            \begin{tabular}[r]{@{}r@{}}911.5\\ {\small (476.7)}\end{tabular} & 
            \begin{tabular}[r]{@{}r@{}}\textbf{1072.8}\\ {\small (664.0)}\end{tabular} &
            \begin{tabular}[r]{@{}r@{}}990.4\\ {\small (707.5)}\end{tabular}\\
 \bottomrule
\end{tabular}
}
\caption{
Average rewards and standard deviations (numbers in bracket) over the last 100,000 time steps over Atari environments.
}
\label{tab:atari_result_std}
\end{table*}
%%%%%%%%%%%%%%%%%%%%%%%%%%%%%%%%%%%%%%%%%%%%%%%%%%%%%%%%%%%%%%%%%%%%%%%%%%%%%%%

\begin{figure}[htp!]
    \includegraphics[clip, trim=0.0cm 0.0cm 0.0cm 0.0cm, width=\textwidth]{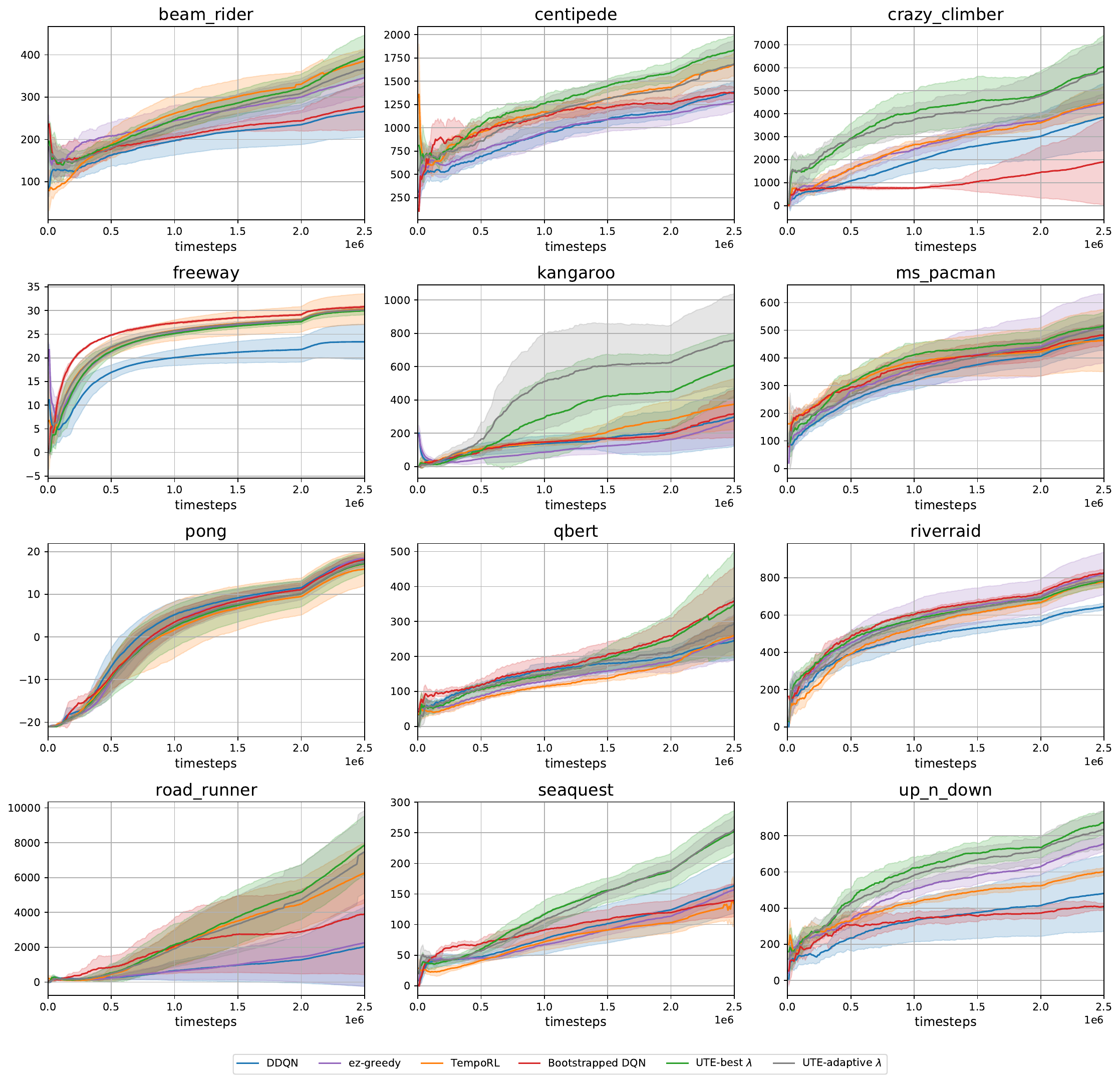}
    \centering
    \caption{Per-game Atari learning curves for \TEE{} with the best $\mu$ for zeta distribution, \Algname{} with the best uncertainty parameter, \Algname{} with adaptive uncertainty parameter and the other baselines.(7 random seeds)}
    \label{fig:atari_learning_curve}
\end{figure}

%%%%%%%%%%%%%%%%%%%%%%%%% DDPG %%%%%%%%%%%%%%%%%%%%%%%%%%%%%
%
\begin{figure}[h!]
    \centering
    \begin{subfigure}[b]{1.0\textwidth}
    \caption{Learning Curve}
        \begin{subfigure}[b]{0.24\textwidth}
            \includegraphics[clip, trim=0.9cm 0.0cm 0.9cm 0.5cm, width=\textwidth]{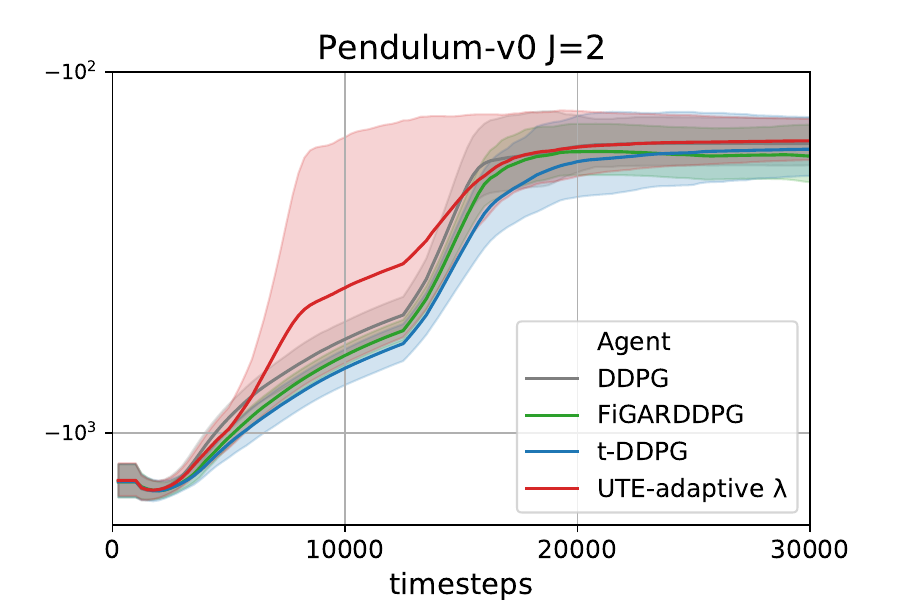}
            \label{fig:ddpg_sub1}
        \end{subfigure}
        \begin{subfigure}[b]{0.24\textwidth}
            \includegraphics[clip, trim=0.9cm 0.0cm 0.9cm 0.5cm, width=\textwidth]{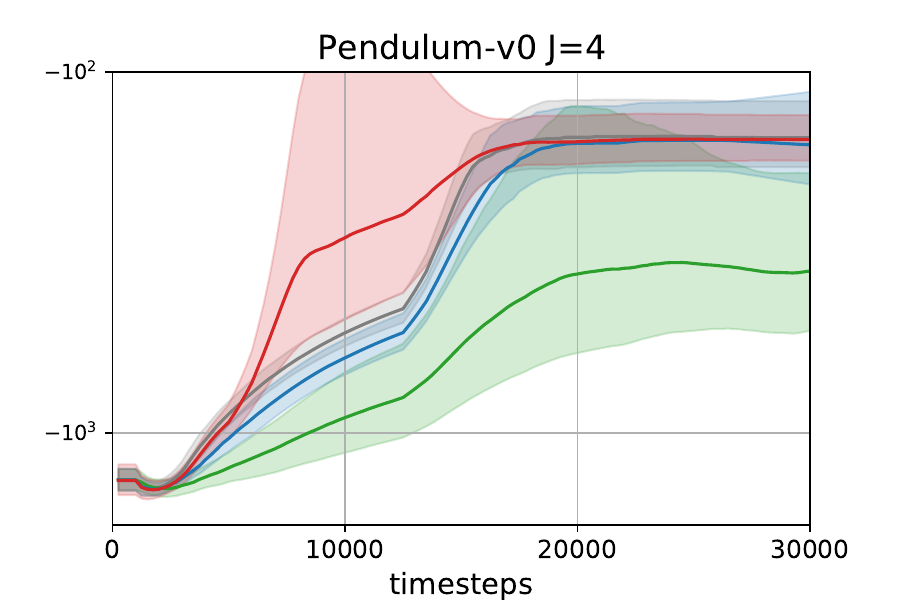}
            \label{fig:ddpg_sub2}
        \end{subfigure}
        \begin{subfigure}[b]{ 0.24\textwidth}
            \includegraphics[clip, trim=0.9cm 0.0cm 0.9cm 0.5cm,width=\textwidth]{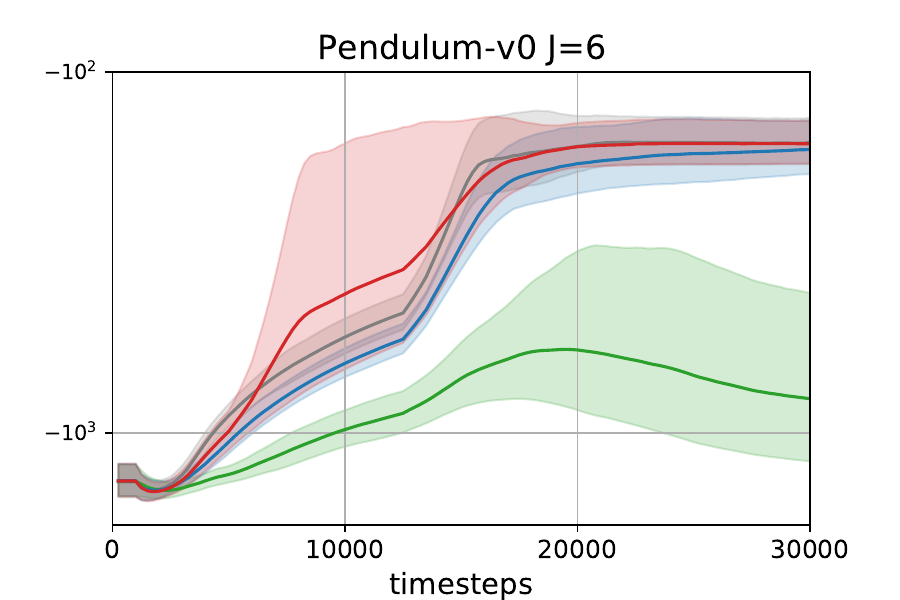}
            \label{fig:ddpg_sub3}
        \end{subfigure}
        \begin{subfigure}[b]{0.24\textwidth}
            \includegraphics[clip, trim=0.9cm 0.0cm 0.9cm 0.5cm, width=\textwidth]{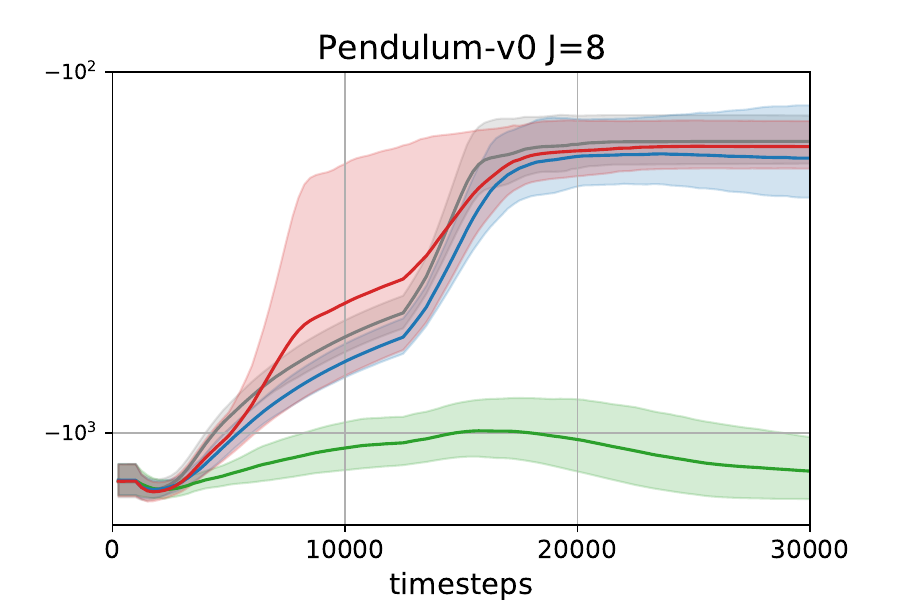}
            \label{fig:ddpg_sub4}
        \end{subfigure}  
        
    \end{subfigure}

    \begin{subfigure}[b]{1.0\textwidth}
    \caption{Number of Decisions per Episode}
        \begin{subfigure}[b]{0.24\textwidth}
            \includegraphics[clip, trim=0.9cm 0.0cm 0.9cm 0.5cm, width=\textwidth]{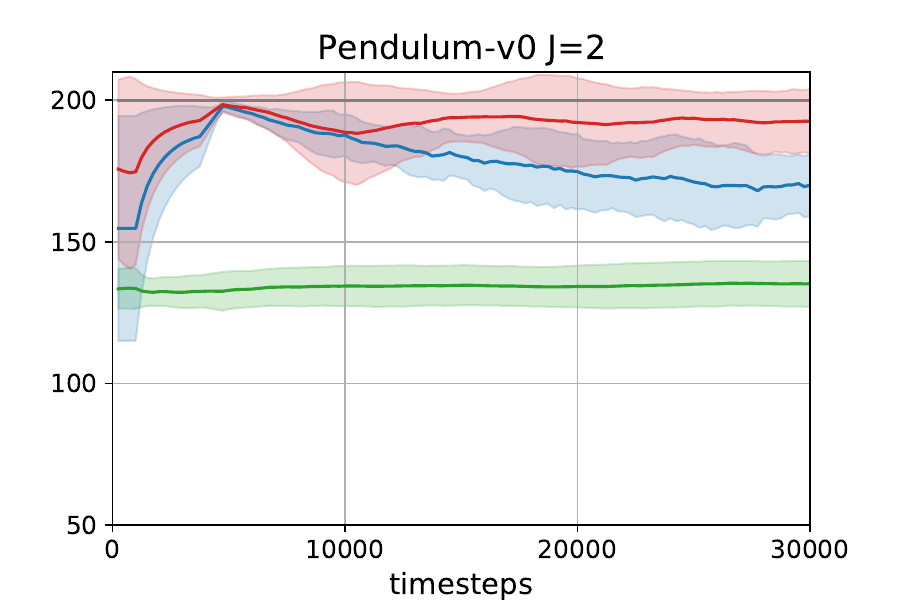}
            \label{fig:ddpg_sub5}
        \end{subfigure}
        \begin{subfigure}[b]{0.24\textwidth}
            \includegraphics[clip, trim=0.9cm 0.0cm 0.9cm 0.5cm, width=\textwidth]{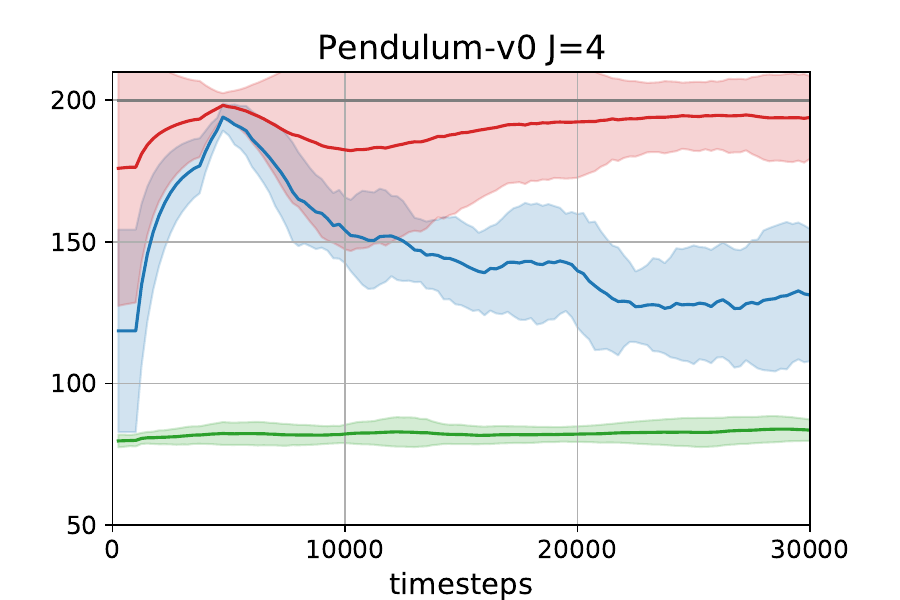}
            \label{fig:ddpg_sub6}
        \end{subfigure}
        \begin{subfigure}[b]{ 0.24\textwidth}
            \includegraphics[clip, trim=0.9cm 0.0cm 0.9cm 0.5cm,width=\textwidth]{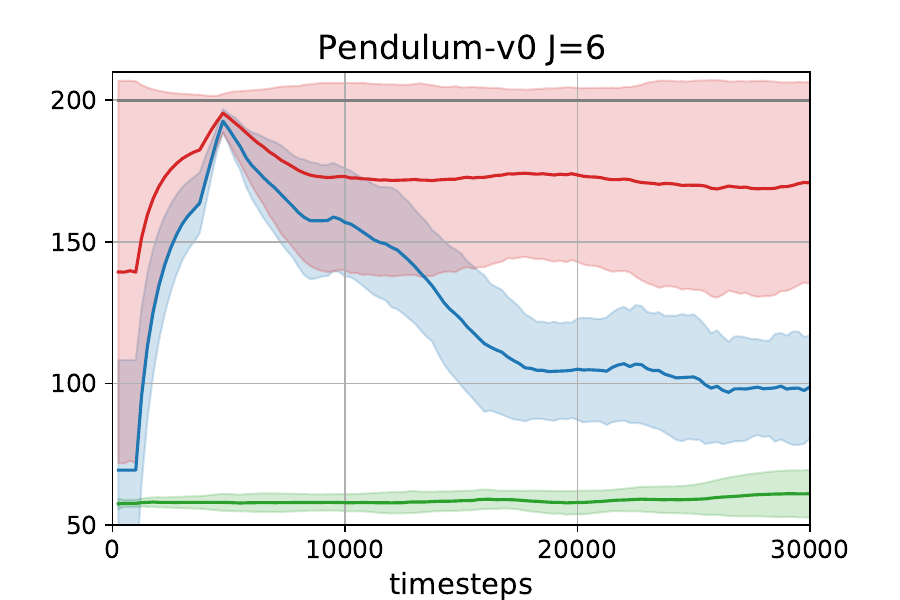}
            \label{fig:ddpg_sub7}
        \end{subfigure}
        \begin{subfigure}[b]{0.24\textwidth}
            \includegraphics[clip, trim=0.9cm 0.0cm 0.9cm 0.5cm, width=\textwidth]{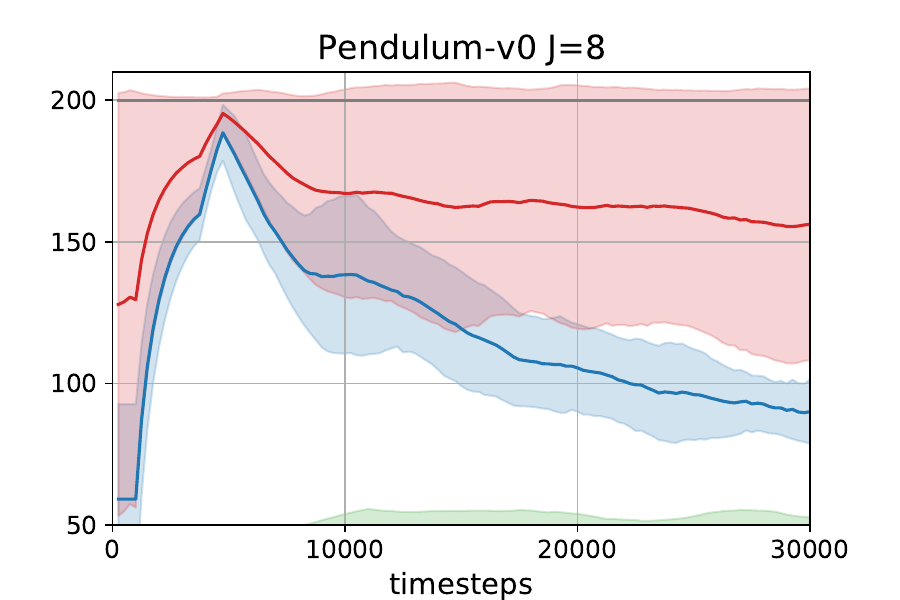}
            \label{fig:ddpg_sub8}
        \end{subfigure}
        
    \end{subfigure}
    
    \caption{Learning curves for \texttt{DDPG}, \texttt{FiGAR}, \texttt{t-DDPG}, \Algname{} with adaptive $\lambda$ on Pendulum-v0. 
    $J$ represents the maximal \rc{} utilized during the training of \Algname{} with adaptive $\lambda$, \texttt{t-DDPG}, and \texttt{FiGAR}. 
    Solid lines represent the average, while the shaded regions reflect the standard deviation across $20$ trials. 
    Images in the top row (a) display the rewards obtained, while those in the bottom row (b) illustrate the steps and decisions needed for each evaluation rollout.}
    \label{fig:Pendulum-v0}
\end{figure}

\subsection{DDPG Implementation Details and Additional Results}
For our base \texttt{DDPG} setup, we used an openly available code from (https://github.com/sfujim/TD3) and maintained its default hyperparameters. 
However, we adjusted the maximum training steps and initial random steps, as detailed in the main paper. 
While we employ a constant epsilon-greedy exploration for the extension policy for \texttt{FiGAR} and \texttt{t-DDPG}~\footnote{We followed the description provided by~\citet{sharma2017learning} and~\citet{biedenkapp2021temporl}}, we refrain from applying epsilon-greedy exploration for \texttt{UTE-DDPG}. 
This is due to our algorithm already integrating a UCB-style exploration strategy for the extension policy.

Regarding our \texttt{UTE-DDPG} configuration, we employ the algorithm as described in Algorithm 1. 
The primary distinction lies in substituting normal $Q$-learning with those specific to \texttt{DDPG} training. 
For instance, in \texttt{DDPG}, the actor's exploration policy involves adding exploration noise instead of adhering to an epsilon-greedy policy. 
Additionally, we again can make use of the base agent's Q-function to learn the option-value function, as described in Equation~\eqref{eq:skip_Qlearning_TempoRL}.
Note that our method is \textbf{generic} so that it can be applied to any other existing algorithms. 
This combination of ease-of-use, significantly improved performance, and wide adaptability not only emphasizes its practicality but also underscores the broad applicability of our approach.

Figure~\ref{fig:Pendulum-v0} depicts the learning curves of various DDPG agents across different maximal \rc{}s. 
The result indicates that \texttt{UTE-DDPG} accelerates learning and achieves superior final rewards compared to other benchmarks, particularly when the maximal \rc{} is small (e.g., $J=2$ or $J=4$).
Furthermore, our approach showcases remarkable stability, with its efficacy largely unaffected by increasing the maximal \rc{}. 
Intriguingly, \texttt{UTE-DDPG} tends to execute action repetitions less frequently than \texttt{t-DDPG}.
In this continuous control environments, recklessly repeating actions can deteriorate the performance significantly.
Thanks to our \textit{uncertainty-aware} extension, actions are repeated carefully, leading to faster learning.

\end{document}